\documentclass[10pt, a4paper,conference]{IEEEtran}

\usepackage[pdftex]{graphicx}
\graphicspath{{./fig/}}
\usepackage{xcolor}
\usepackage{subcaption}
\usepackage{enumitem}

\usepackage{amsmath}
\usepackage{amssymb}
\usepackage{amsfonts}

\begin{document}

\title{Data Augmentation via Mixed Class Interpolation using Cycle-Consistent Generative Adversarial Networks Applied to Cross-Domain Imagery}

\author{\IEEEauthorblockN{Hiroshi Sasaki$^1$, Chris G. Willcocks$^1$, Toby P. Breckon$^{1,2}$}
\IEEEauthorblockA{Department of \{$^1$Computer Science $|$ $^2$Engineering\},
Durham University,
Durham, UK
}

}

\maketitle


\begin{abstract}

Machine learning driven object detection and classification within non-visible imagery has an important role in many fields such as night vision, all-weather surveillance and aviation security. However, such applications often suffer due to the limited quantity and variety of non-visible spectral domain imagery, in contrast to the high data availability of visible-band imagery that readily enables contemporary deep learning driven detection and classification approaches. To address this problem, this paper proposes and evaluates a novel data augmentation approach that leverages the more readily available visible-band imagery via a generative domain transfer model. The model can synthesise large volumes of non-visible domain imagery by image-to-image (I2I) translation from the visible image domain. Furthermore, we show that the generation of interpolated mixed class (non-visible domain) image examples via our novel Conditional CycleGAN Mixup Augmentation (C2GMA) methodology can lead to a significant improvement in the quality of non-visible domain classification tasks that otherwise suffer due to limited data availability. Focusing on classification within the Synthetic Aperture Radar (SAR) domain, our approach is evaluated on a variation of the Statoil/C-CORE Iceberg Classifier Challenge dataset and achieves 75.4\% accuracy, demonstrating a significant improvement when compared against traditional data augmentation strategies (Rotation, Mixup, and MixCycleGAN). 

\end{abstract}




\IEEEpeerreviewmaketitle


\section{Introduction}
The demand for automated pattern recognition, especially automatic object detection and classification in imagery, is continuously expanding. In computer vision, there are many applications utilising automatic pattern recognition, for example, optical character recognition~\cite{casey1996survey}, video surveillance~\cite{hu2004survey}, agricultural analysis from satellite imagery~\cite{vibhute2012applications}, and defect detection in factory automation~\cite{huang2015automated}. These functions are enabled by recent advances in machine learning, namely deep neural networks (DNN)~\cite{goodfellow2015deep}. DNN have enabled hitherto unprecedented performance on various challenging computer vision tasks such as image classification, object detection, semantic segmentation and temporal video analysis.


\begin{figure}[tb]
  \begin{center}
    \includegraphics[clip,width=9.0cm]{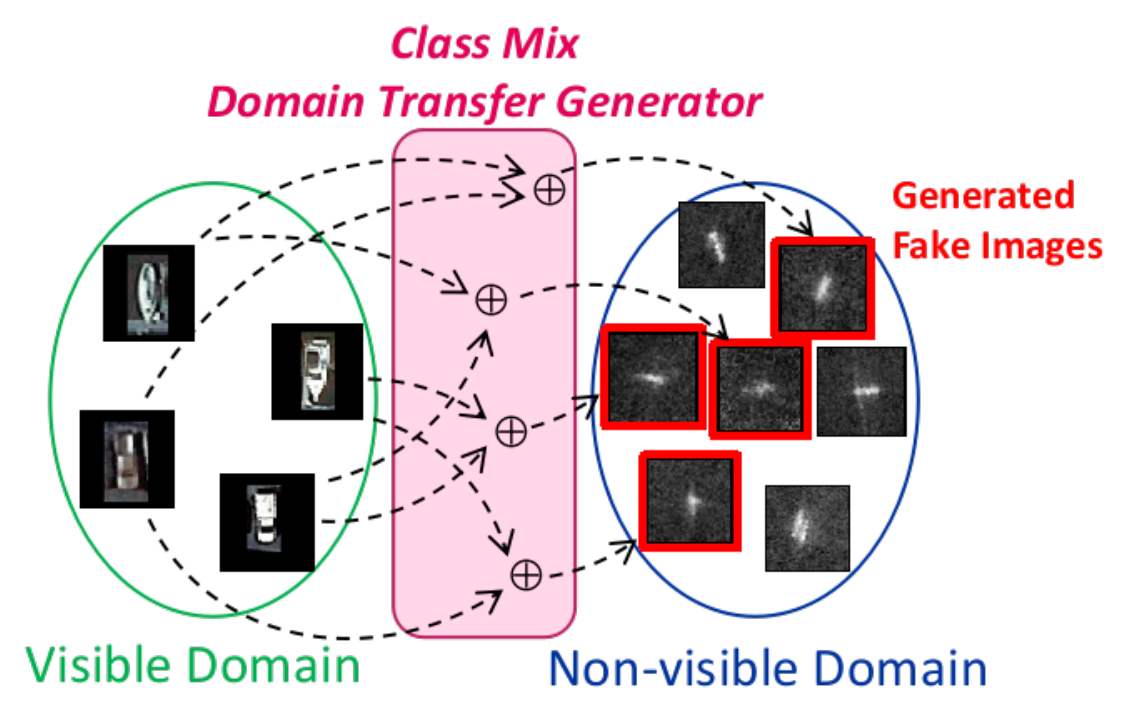}
    \caption{Conceptual illustration of our novel data augmentation approach for generating cross-domain, class-interpolated image instances.}
    \label{fig:overall}
  \end{center}
\end{figure}


This expansion, both in demand and performance, has led to the broader consideration of computer vision applications in imagery domains beyond the visible spectrum, i.e. non-visible images such as infrared (thermal)~\cite{kundegorski2016real}, synthetic aperture radar (SAR)~\cite{chen2016target} and X-ray images~\cite{akcay2018using}. Imaging within the non-visible spectrum provides sensing capabilities ranging from all-weather visibility, object temperature, material characteristics and sub-surface/object transparency. Whilst DNN approaches have predominately been applied to visible domain imagery, they are readily applied across the non-visible spectrum. However, the primary challenge is the low data availability in these additional spectral imaging domains. Whilst contemporary DNN approaches generally perform well in domains with large amounts of data available, within the non-visible imaging domain data availability is often more limited and it can be difficult to collect enough image samples to provide sufficient variability and coverage of the target data distribution expected at inference (test, deployment) time. For example, SAR imagery is far less readily available and accessible due to both the lesser prevalence of this sensing technology and its associated costs. In addition, SAR imagery substantially differs from visible-band imagery because it results from active sensing by microwave radar backscatter projection, whilst visible images are captured passively according to the intensity of reflected scene illumination. Moreover, SAR imagery is significantly impacted by the choice of microwave bands in use and by the angle of microwave transmission. These variations from conventional imagery that preclude the direct applicability of commonplace transfer learning solutions, coupled with the lack of data availability, further inhibit inter-task applications with such diverse sensor imagery.

In order to address this issue of DNN model generalisation under such limited data availability, data augmentation methods such as geometric image transformation and pixel-wise intensity transformations are traditionally adopted. However, such methods tend to synthesise images which are highly biased to both the prior assumptions of this augmentation and the prior distribution of the already limited dataset in use. An alternative solution, more specific to object classification tasks, involves blending a pair of input images of different classes to smooth the classification decision boundary during the training \cite{zhang2018mixup}. This approach can be effective when there are few training examples (limited data availability), but remains highly sensitive to biases in the input samples. 
To overcome these issues, recent research into image synthesis and dataset augmentation has focused on stochastic generative models, which can create a variety of high-quality images~\cite{goodfellow2014generative}. In particular, image-to-image (I2I) translation models are able to generate samples by mapping between image domains~\cite{zhu2017unpaired}, whereas standard generative models synthesise images by transforming random values sampled from a simpler prior distribution. I2I translation is particularly effective when there are few images in a desired domain and large quantities of data available in another indirectly related domain, such as in the context of SAR images and publicly available visible images. 

Taking this into consideration, we exploit the potential of I2I translation as a dataset augmentation strategy and develop a new I2I translation model, adopted from Cycle-Consistent Generative Adversarial Networks (CycleGAN)~\cite{zhu2017unpaired}. In particular, we modify CycleGAN by manipulating class conditional information and generating class-interpolated images~(Figure \ref{fig:overall}), as described in detail in Section~\ref{sec:method}. The experiments supporting our method, within the context of SAR object classification, are presented in Section~\ref{sec:experiments} with subsequent conclusions presented in Section~\ref{sec:conclusion}.


\section{Related Work}
Many data augmentation approaches within a computer vision context have been proposed and divided into two sub-types: unsupervised and supervised \cite{shijie2017research}.

\subsection{Unsupervised Data Augmentation}
An unsupervised approach aims to increase the quantity of training imagery via a set of fixed geometric and pixel-wise image processing 
operations to transform an existing dataset image (e.g. flipping, rotation, cropping, adding noise, etc. ~\cite{shijie2017research}).

Mixup~\cite{zhang2018mixup} is a recent approach that blends pairs of randomly chosen training images using randomly weighted blending rates to avoid overfitting. In addition,~\cite{zhong2017random}~\cite{terrance2017improved}~\cite{chen2020gridmask}~\cite{ghiasi2018dropblock} have shown the effectiveness of partially masking image sub-regions to force generalisation during model training. Instead of zero maskings, CutMix~\cite{yun2019cutmix} replaces these regions with a region of the same size from another training set image and provides an improvement in performance.

\subsection{Supervised Data Augmentation}
While unsupervised methods can reduce overfitting, the trained models are often unable to accurately model patterns or trends that appear within the test distribution that are infrequent within the training data distribution.
This is largely due to the fact that unsupervised augmentation approaches transform data sampled from the same underlying training distribution, therefore their outputs reflect the inherent biases and patterns in this original training distribution.
In order to overcome this issue, several supervised approaches have been proposed that instead generate new images using additional label information to improve generalisation between domains~\cite{verma2019manifold}~\cite{philip2019style}.

Manifold Mixup~\cite{verma2019manifold} is a modification of Mixup. This interpolates not only input images and their associated output labels but also latent information within the hidden layers. This attempts to increases the novelty of data samples generated by latent information level processing. Meanwhile, 
data augmentation via diversification of image style was proposed~\cite{philip2019style}. Utilising a style transfer network~\cite{golnaz2017exploring}, a DNN trained to transfer the style from one image to another while preserving its semantic content, they additionally augmented their training data via image style randomisation.

Generative Adversarial Networks (GAN)~\cite{goodfellow2014generative} have significantly impacted data augmentation within DNN training. A GAN is a generative DNN architecture, designed to have a generator and a discriminator component that compete against each other during its training process. The generator is trained to map randomised values to real data examples by the discriminator output. The discriminator is simultaneously trained to discriminate real and fake data examples produced by the generator. The objective function is defined as:
\begin{eqnarray}
  \min_{G}\max_{D}V(D,G) = \mathbb{E}_{x \sim p_{\text{data}(x)}}[\log D(x)] + \nonumber \\
  \mathbb{E}_{z \sim p_{x}(z)}[\log (1-D(G(z)))] \label{eqn:gan}
\end{eqnarray}
where $G$ and $D$ are the generator and discriminator respectively. $x$ is input data and $z$ is random noise.
As a result of training the generator within this GAN architecture, it is hence optimised to create realistic, yet artificial data that is statistically similar (drawn from the same distribution) as the real data. The GAN architecture has been shown to be effective with convolutional neural networks, popularised by Deep Convolutional GAN (DCGAN)~\cite{radford2016unsupervised}.
A Conditional GAN (cGAN)~\cite{mirza2014conditional} was proposed to modify the GAN architecture to take account of classes by adding class labels into the inputs of the generator and discriminator. The objective function~(\ref{eqn:gan}) is modified as:
\begin{eqnarray}
  \min_{G}\max_{D}V(D,G) = \mathbb{E}_{x \sim p_{\text{data}(x)}}[\log D(x|y)] + \nonumber \\
  \mathbb{E}_{z \sim p_{x}(z)}[\log (1-D(G(z|y)|y))]
\end{eqnarray}
where $y$ is the category label given in the objective function. Moreover, another GAN variant, called Auxiliary Classifier GAN (ACGAN)~\cite{odena2017conditional}, implemented classification in addition to generative modelling. This architecture trains its network to minimise the distance of between both the real and fake data examples and the actual and predicted category labels. 
While such conditional information was initially implemented as a concatenation of the input and output of the networks, other methods find that the conditional information can be incorporated into the normalisation layers to significantly improve the generated results~\cite{dumoulin2017learned}~\cite{park2019semantic}.
These normalisation layers are called the conditional normalisation layers and the generators are modified as $G(z,e(y))$, where $e$ is the embedding function. 
The effectiveness of this conditional label embedding has been not only been used in the generator, but also to the discriminator.  This `projection discriminator' is implemented by an inner product of the embedded one-hot labels and the intermediate layer outputs~\cite{miyato2018cgans}. 

A large corpus of images from other related domains can also be useful for increasing training data in some cases. Generating new images by transferring from another domain image set, which is called I2I translation, has the possibility of expanding the distribution of training data such that it retains more of the structure of real images rather than the synthesised images generated only from vectors of noise. CycleGAN~\cite{zhu2017unpaired} is one of the expansions of GAN specified in I2I translation. In this method, $G$ and $D$ are trained to transfer from source images $x_s \in X_s$ to target images $x_t \in X_t$. This not only learns a lateral transform, but also the bilateral transform paths $G_t(x_s),G_s(x_t)$. In addition, this adopts a new loss measure named a cycle-consistency loss $L_{\text{cyc}}(G_s,G_t)$, which is represented as:
\begin{eqnarray}
 L_{\text{cyc}}(G_s,G_t) = \mathbb{E}_{x_s \in X_s}[\|G_s(G_t(x_s))-x_s\|_1] + \nonumber \\
 \mathbb{E}_{x_t \in X_t}[\|G_t(G_s(x_t))-x_t\|_1]
\end{eqnarray}
In total, the full objective function is:
\begin{eqnarray}
 \min_{G_s,G_t}\max_{D_s,D_t} V(D_s, G_s) + V(D_t, G_t) + \lambda_\text{cyc} L_{\text{cyc}}(G_s,G_t)
\end{eqnarray}
where $\lambda_\text{cyc}$ is a cycle-consistency loss weight.

MixCycleGAN~\cite{liang2018understanding} applies a `mixup' operation to the CycleGAN process to stabilise the training and increase the variety of the generated outputs. This method splits an input image into two rectangular regions vertically or horizontally and replaces one region with that of another image:
\begin{eqnarray}
 \bar{x} = cat(x_1[:\lambda H, :], x_2[(1-\lambda)H:, :]) \nonumber \\
 \text{or}~cat(x_1[:, :\lambda W], x_2[:, (1-\lambda)W:])
\end{eqnarray}
where, $\bar{x}$ is the mixed image, $x_1, x_2 \in X$ are the input images, $H, W$ are the height and width of the input images respectively, and $cat$ is a concatenation function. $\lambda \in [ 0, 1 ]$ is the mixup ratio, and $\lambda \sim \text{Beta}(\alpha,\alpha)$ is from the beta distribution $\text{Beta}$, in which $\alpha$ is constantly set as in \cite{zhang2018mixup}. The preprocessed mixed image $\bar{x}$ is input to the generator $G$ of CycleGAN to synthesise a fake image. The discriminator $D$ is modified to estimate the mixup ratio from the alpha-blended real and fake images, which is optimised as:
\begin{eqnarray}
  \min \mathbb{E}_{x \in X}[\log | \lambda-D(\lambda x+(1-\lambda)G(\bar{x})) |]
  \label{eqn:mixup_d}
\end{eqnarray}

Our approach is similar to MixCycleGAN. However, while MixCycleGAN stitches rectangular image regions and does not use class labels, our approach adopts cGAN with the conditional normalisation layers and the projection discriminator to allow the class labels as input for the generator to enforce synthesising class-specific images. This proposed strategy enables generation of more sophisticated class-interpolated images by alpha-blending of the input images and class labels, rather than with a simple rectangular image region mixup. The details of these are described in Section~\ref{sec:method}.


\section{Methodology}
\label{sec:method}
The proposed method assumes a source domain dataset $(x_s^i,y_s^i) \in X_{s}^N$ and a target domain dataset $(x_t^j,y_t^j) \in X_{t}^M$ which consist of $N$ and $M(\ll N)$ samples respectively. $x_s^i$ and $x_t^j$ are the images themselves and $y_s^i$ and $y_t^j$ are class labels. The types of classes are common in both domains.
\begin{figure}[tb]
  \begin{center}
    \includegraphics[clip,width=8.5cm]{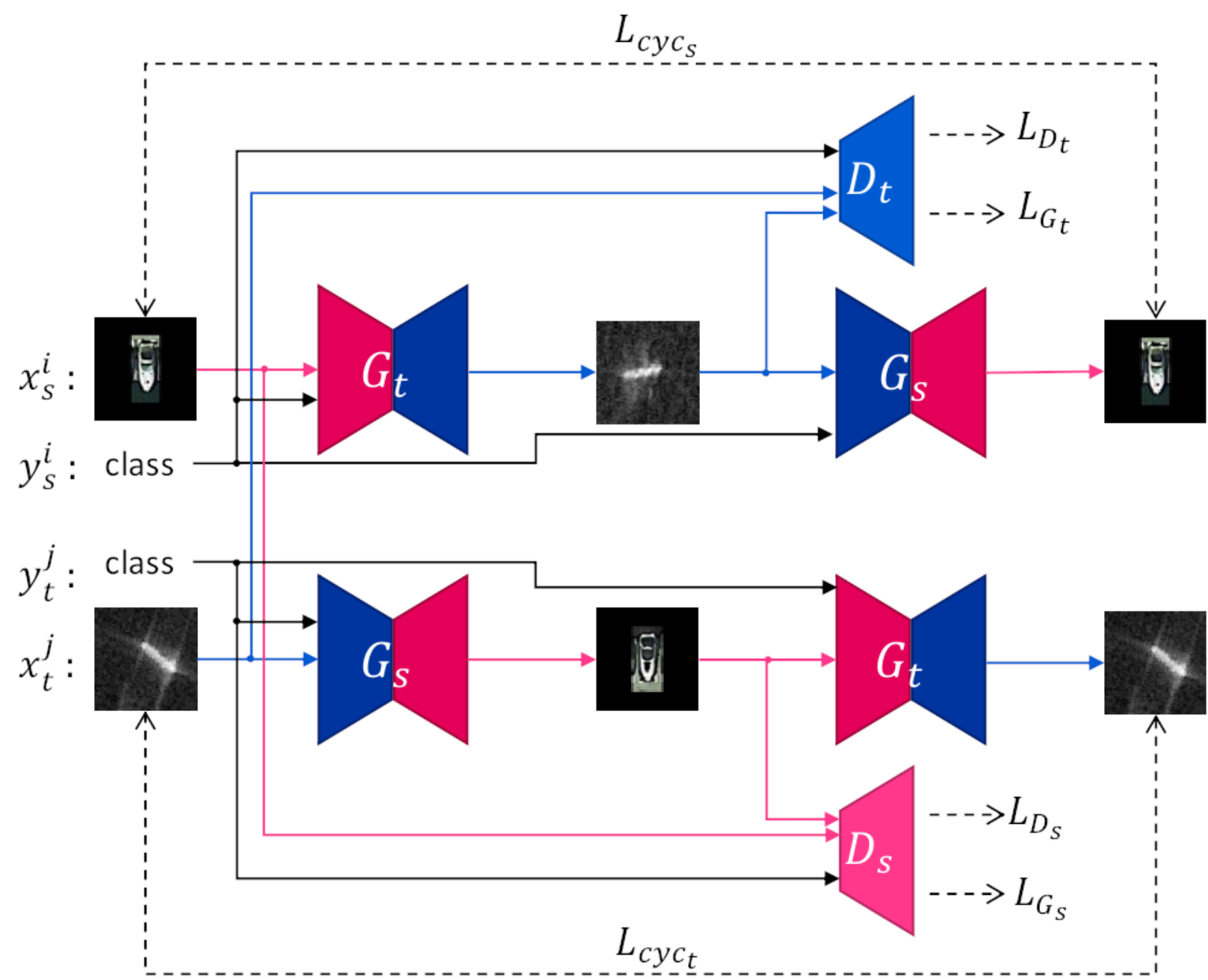}
    \caption{Overall flow of our conditional CycleGAN model.}
    \label{fig:model}
  \end{center}
\end{figure}

Initially, a generative model, which transfers between two different domains, is built using the conditional CycleGAN approach. In order to prevent mode collapse and stabilise training, Spectral Normalization~\cite{miyato2018spectral} is combined with the gradient penalty~\cite{gulrajani2017improved} as proposed in \cite{chu2020Smoothness}. Furthermore, as discussed previously, we apply conditional regularisation of cGAN to our CycleGAN model by implementing conditional normalisation layers and projection discriminators to improve the output quality. The overall flow is shown in (Figure~\ref{fig:model}) where, unlike ordinary CycleGAN, the generator and discriminator functions are conditioned on the class labels.
The objective function is defined as a simple sum of weighted terms:
\begin{eqnarray}
    L = \lambda_{s}L_{G_{s}} + \lambda_{t}L_{G_{t}} + \lambda_{s}L_{D_{s}} + \lambda_{t}L_{D_{t}} + \nonumber \\
    \lambda_{s}\lambda_{\text{cyc}}L_{\text{cyc}_{s}} + \lambda_{t}\lambda_{\text{cyc}}L_{\text{cyc}_{t}}
\end{eqnarray}
where:
\begin{eqnarray}
  && L_{G_{s}} = \nonumber \\
  && \quad \mathbb{E}_{(x_t^j,y_t^j) \in X_{t}}[\log (1\mathalpha{-}D_{s}(G_{s}(x_t^j,e_{t}(y_t^j)),e_{s}(y_t^j)))] \\
  && L_{G_{t}} = \nonumber \\
  && \quad \mathbb{E}_{(x_s^i,y_s^i) \in X_{s}}[\log (1\mathalpha{-}D_{t}(G_{t}(x_s^i,e_{s}(y_s^i)),e_{t}(y_s^i)))] \\
  && L_{D_{s}} = \mathbb{E}_{(x_s^i,y_s^i) \in X_{s}}[\log (1-D_{s}(x_s^i,e_{s}(y_s^i)))] + \nonumber \\
  && \qquad \mathbb{E}_{(x_t^j,y_t^j) \in X_{t}}[\log (D_{s}(G_{s}(x_t^j,e_{t}(y_t^j)), e_{s}(y_t^j)))] + \nonumber \\
  && \qquad \lambda_{\text{gp}}\mathbb{E}_{(\hat{x}_s^j,\hat{y}_s^j) \sim \mathbb{P}_{\hat{x}_s,\hat{y}_s}}[(\| \nabla D_{s}(\hat{x}_s^j, e_{s}(\hat{y}_s^j))\|_2-1)] \\
  && L_{D_{t}} = \mathbb{E}_{(x_t^j,y_t^j) \in X_{t}}[\log (1-D_{t}(x_t^j,e_{t}(y_t^j)))] + \nonumber \\
  && \qquad \mathbb{E}_{(x_s^i,y_s^i) \in X_{s}}[\log (D_{t}(G_{t}(x_s^i,e_{s}(y_s^i)), e_{t}(y_s^i)))] + \nonumber \\
  && \qquad \lambda_{\text{gp}}\mathbb{E}_{(\hat{x}_t^j,\hat{y}_t^j) \sim \mathbb{P}_{\hat{x}_t,\hat{y}_t}}[(\| \nabla D_{t}(\hat{x}_t^j, e_{t}(\hat{y}_t^j))\|_2-1)] \label{eqn:loss_d_t}\\
  && L_{\text{cyc}_{s}} = \nonumber \\
  && \mathbb{E}_{(x_s^i,y_s^i) \in X_{s}}[\| (G_{s}\!(G_{t}\!(x_s^i,e_{s}(y_s^i)),e_{t}(y_s^i)) \mathalpha{-} x_s^i)\|_1] \\
  && L_{\text{cyc}_{t}} = \nonumber \\
  && \mathbb{E}_{(x_t^j,y_t^j) \in X_{t}}[\| (G_{t}\!(G_{s}\!(x_t^j,e_{t}(y_t^j)),e_{s}(y_t^j)) \mathalpha{-} x_t^j)\|_1]
\end{eqnarray}
$\lambda_{s}$ and $\lambda_{t}$ are source domain and target domain weights, respectively. $\lambda_{\text{gp}}$ is a weight of the gradient penalty. That is, we balance the corresponding generator and discriminator functions with the cycle-consistency losses for both the source and target domains accordingly.

After training, the model is used for
the synthesis of new class-conditioned images via the domain transfer. A pair of images and class labels in the source domain dataset $(x_{s}^i,y_{s}^i),(x_{s}^j,y_{s}^j) \in X_{s}^N$ are used as an input.
Subsequently, the input is processed to produce a tuple of a mixed image, label, and embedded feature vector $(\bar{x}_s^k, \bar{y}_s^k, \bar{e}_s^k)$, defined by:
\begin{eqnarray}
  \bar{x}_s^k = x_{s}^i * \lambda + x_{s}^j * (1-\lambda) \\
  \bar{y}_s^k = y_{s}^i * \lambda + y_{s}^j * (1-\lambda) \\
  \bar{e}_s^k = e_{s}(y_{s}^i) * \lambda + e_{s}(y_{s}^j) * (1-\lambda)
\end{eqnarray}
where $\lambda \in [ 0, 1 ]$ is the mixup ratio, and $\lambda \sim \text{Beta}(\alpha,\alpha)$ from the beta distribution $\text{Beta}$, in which $\alpha$ is constantly set as in \cite{zhang2018mixup}. As a result, the mixed pair $(\tilde{x}_t^k, \tilde{y}_t^k)$ that is input to the generator and discriminator is defined, where:
\begin{eqnarray}
  (\tilde{x}_t^k, \tilde{y}_t^k) = (G_{t}(\bar{x}_s^k,\bar{e}_s^k), \bar{y}_s^k)
\end{eqnarray}
As a result, $N$ samples $\tilde{X}_{t}^N=\{(\tilde{x}_t^k, \tilde{y}_t^k)\}$ are synthesised. The new fake samples are combined with the original dataset as \mbox{$X_{t}^M \cup \tilde{X}_{t}^N$}, where we denote this method as Conditional CycleGAN Mixup Augmentation (C2GMA).


\section{Experiments}
\label{sec:experiments}
The method is evaluated in the context of the ships/icebergs SAR classification task using the Statoil/C-CORE Iceberg Classifier Challenge dataset~\cite{statoil-c-core}. Results are compared between classification models trained with and without existing dataset augmentation approaches in addition to our proposed CycleGAN driven C2GMA (Section \ref{sec:method}) approaches.

\subsection{Dataset}
\begin{figure}[tb]
  \begin{center}
    \includegraphics[clip,width=8.5cm]{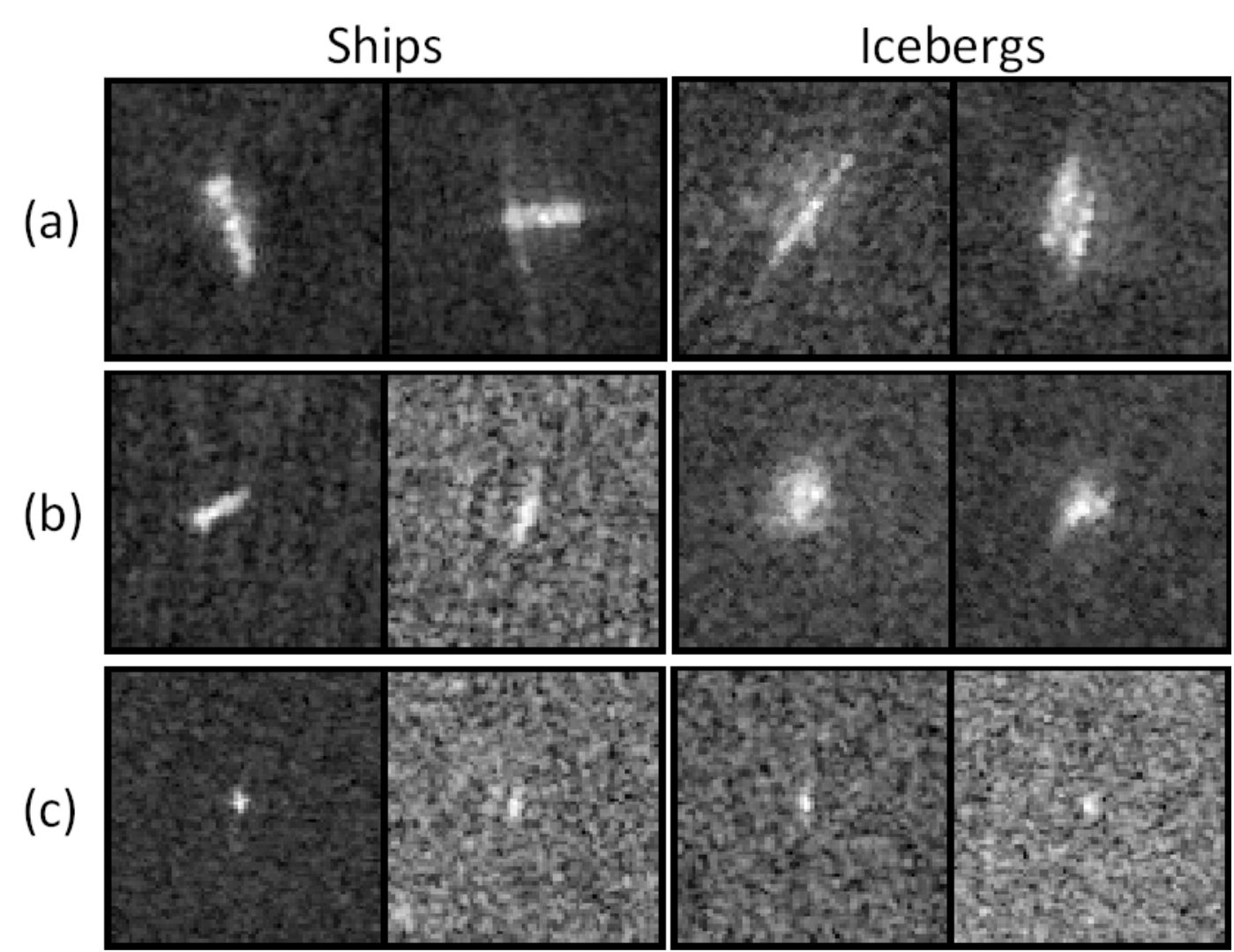}
    \caption{SAR ships/icebergs images divided into three groups based on difficulty of discrimination by distance, angle, object size, etc.}
    \label{fig:sar_ship_iceberg}
  \end{center}
\end{figure}
The Statoil/C-CORE Iceberg Classifier Challenge dataset \cite{statoil-c-core} has a collection of satellite SAR images of ships and icebergs, each with $75\times 75$ pixels. The dataset comprises of a training set with images labelled as either a ship or an iceberg, alongside a set of unlabelled test images. We use only the labelled training data in our experiments (we split this labelled data into different groups for evaluation, discussed subsequently). Each sample in the data is represented by 2-channel floating-point images according to the two different channels of microwave echos: HH and HV. The values in the HH channel are the intensity of the horizontally echos of the horizontal transmitted microwave, whereas the HV channel is the intensity of the vertical echos of the same transmitted microwave.

A challenge of assessing the generalisation performance, given a dataset sampled from a single distribution, is that it does not reflect the case where the distribution of data under the expected testing conditions differs from the distribution of data sampled for training. Therefore, we split the dataset into three groups of discriminable classes, from which the images are sampled at different ratios between training and testing. We initially combine the two channels into one channel:
\begin{eqnarray}
  \label{eq:sar_img}
  I(x,y) = \sqrt{I_{\text{HH}}(x,y)^2 + I_{\text{HV}}(x,y)^2}
\end{eqnarray}
where $I(x,y)$, $I_{\text{HH}}(x,y)$, and $I_{\text{HV}}(x,y)$ are the pixel values of the combined image, the HH image, and the HV image at $(x,y)$ respectively. The dataset is then subdivided into three groups by hand for each class: (a) easily discriminable sets, (b) moderately discriminable sets, and (c) difficult cases (Figure~\ref{fig:sar_ship_iceberg}).

Each of the groups is partitioned into training and testing splits and subsampled at different ratios, where specifically we distort the distribution of the training sets to simulate further imbalance and mismatch between the training distribution and the expected testing data distribution. These splits, and the corresponding skewed subsamplings, are shown in  Table~\ref{tab:dataset}.
\begin{table}[tb]
  \begin{center}
    \caption{The number of samples in the experiment dataset separated by the test set and the three different training sets. The columns (a), (b), and (c) represent: easily identifiable samples, moderate samples, and difficult samples.}
    \begin{tabular}{l r r r r r r r r r} \hline
       & \multicolumn{4}{c}{Ship} && \multicolumn{4}{c}{Iceberg} \\ \cline{2-5} \cline{7-10}
       & (a) & (b) & (c) & total && (a) & (b) & (c) & total \\ \hline \hline
      Test & 97 & 158 & 171 & 426 && 99 & 137 & 141 & 377 \\
      Train \#1 & 96 & 15 & 17 & 128 && 99 & 13 & 14 & 126 \\
      Train \#2 & 96 & 15 & 17 & 128 && 9 & 137 & 14 & 160 \\
      Train \#3 & 96 & 15 & 17 & 128 && 9 & 13 & 140 & 162 \\ \hline
    \end{tabular}
    \label{tab:dataset}
  \end{center}
\end{table}

In order to augment the training datasets using our proposed method, we use the satellite visible image dataset named DOTA \cite{xia2018dota}, which is a collection of commercial satellite images containing many objects such as vehicles annotated with bounding boxes and class labels. Therefore we use visible and SAR image pairs with SAR images originating from the Statoil/C-CORE Iceberg Classifier Challenge dataset~\cite{statoil-c-core} and visible images from the DOTA \cite{xia2018dota} dataset. Due to the lack of iceberg visible images within either dataset, we pair iceberg SAR images from the Statoil/C-CORE Iceberg Classifier Challenge dataset~\cite{statoil-c-core} with representative non-ship images from the DOTA \cite{xia2018dota} dataset, for which purposes we use visible images of vehicles. Despite this obvious semantic mismatch in the second pairing, our I2I translation model specifically synthesises images conforming to the true distribution of the SAR iceberg images as enforced by the discriminator criteria of the loss function in Equation~(\ref{eqn:loss_d_t}).
\begin{figure}[tb]
  \begin{center}
  \centering
    \includegraphics[clip,width=8.5cm]{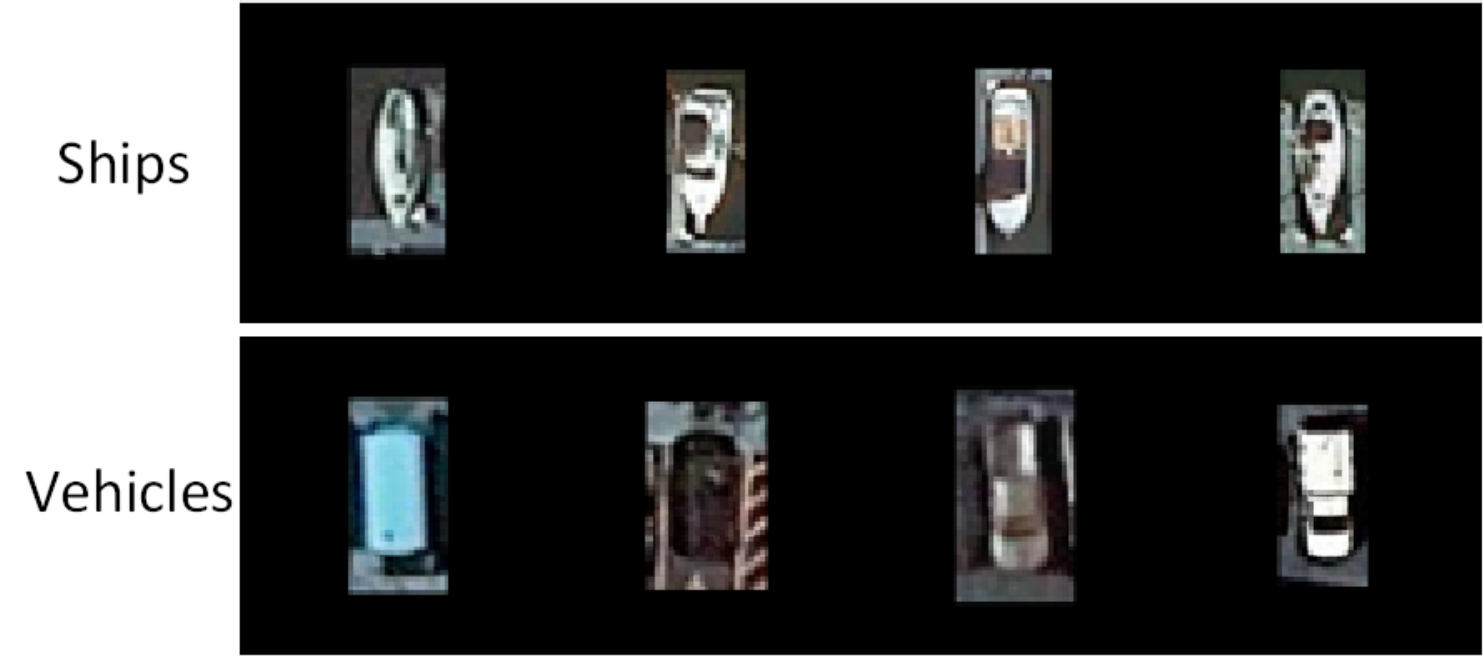}
    \caption{Visible images from~\cite{xia2018dota} (domain transfer source).}
    \label{fig:dota}
  \end{center}
\end{figure}
\begin{figure}[tb]
  \begin{center}
  \centering
    \includegraphics[clip,width=8.0cm]{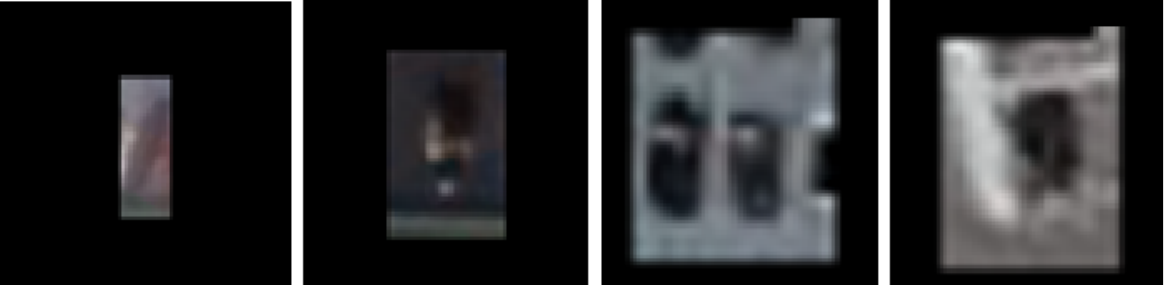}
    \caption{Poor quality visible images illustrating blurriness and multiple objects (which we eliminate).}
    \label{fig:kicked_dota}
  \end{center}
\end{figure}

Initially, visible object images are extracted from the visible dataset using the annotations. Each extracted image is resized in the same way as the SAR image, and its rotations adjusted accordingly. The backgrounds are set to black, which prevents including surrounding objects, which would be undesirable (Figure~\ref{fig:dota}).
The source domain visible dataset exhibits several images that are unclear or incorrect, as in Figure~\ref{fig:kicked_dota}. Such images are eliminated based on their distances from the median of all of the images within each class. These distances are measured in the latent spaces trained by a Variational Autoencoder~\cite{pu2016variational} on individual classes. Using the encoder, all of the images are embedded in a lower dimensional latent space that follows an approximate normal distribution, and the distances of each sample $d(x_{i}^c)$ are calculated:
\begin{eqnarray}
  d(x_{i}^c) = \sqrt{(f_{e}^c(x_{i}^c)-\mathcal{M}^c)^T {S^c}^{-1} (f_{e}(x_{i}^c)-\mathcal{M}^c)} \\
  S^c = \mathbb{E}[(f_{e}(x_{i}^c)-\mathcal{M}^c)(f_{e}(x_{i}^c)-\mathcal{M}^c)^T]
\end{eqnarray}
where $x_{i}^c$ is the $i-th$ input sample of class $c$, $f_{e}$ is the encoder, and $\mathcal{M}^c$ is the median of the encoded features in class $c$. $S^c$ is a normalisation factor for each dimension of the feature vectors in class $c$.
Half of the shorter distance samples are selected for each class, subsampling 14,034 visible ship images and 13,063 visible vehicles, resulting in clearer data and higher-quality annotations for use as our source domain.
\begin{figure}[tb]
  \begin{center}
    \includegraphics[clip,width=9.0cm]{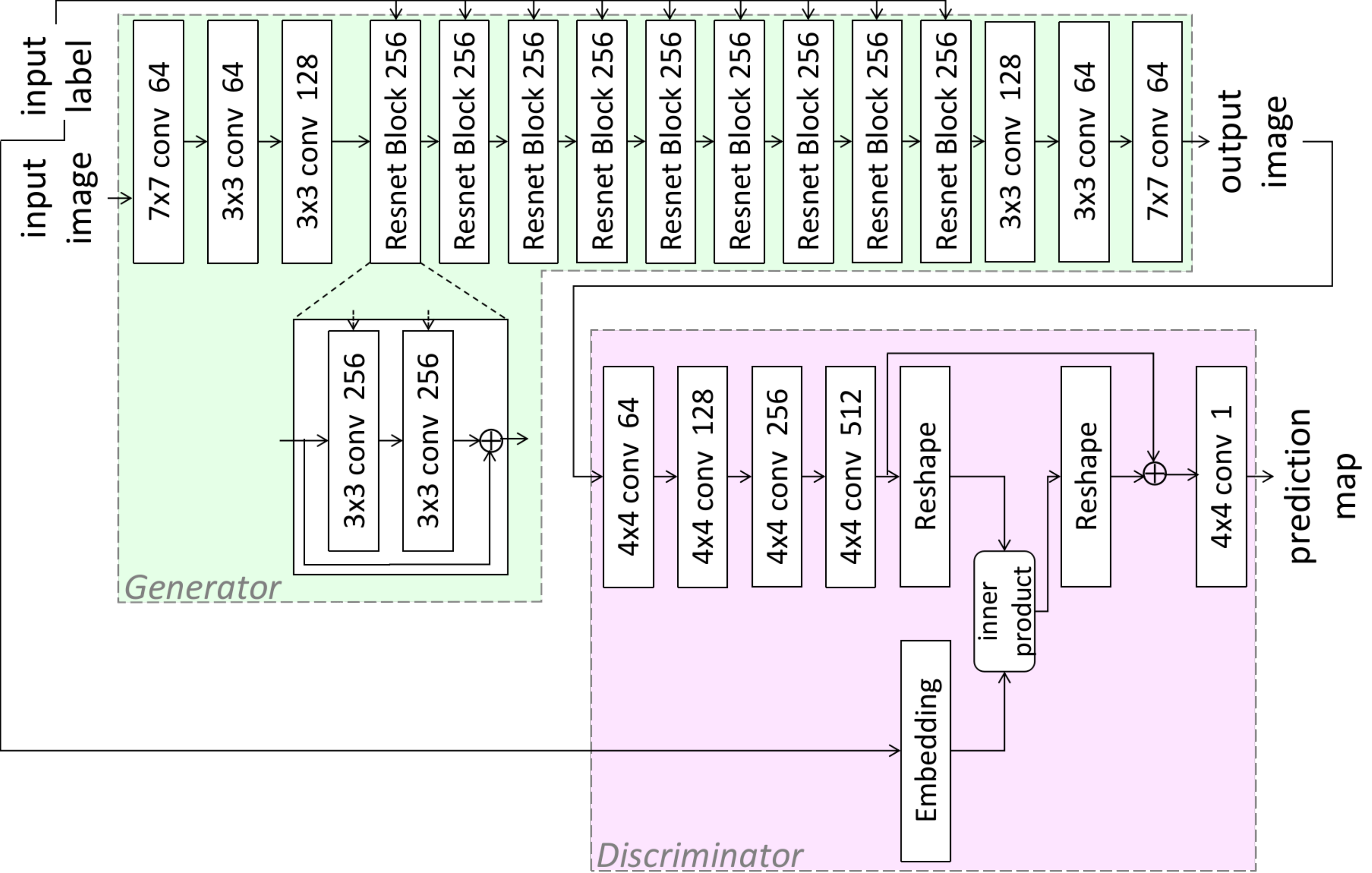}
    \caption{Our network architecture:-  Conditional Batch Normalisation layers are applied to every convolutional layer within the Generator whilst Instance Normalisation layers and Spectral Normalization are applied to every convolutional layer within the Discriminator.}
    \label{fig:net_archs}
  \end{center}
\end{figure}

\vspace{3mm}
\subsection{Training Domain Transfer Model}
Domain transfer models, as described in Section \ref{sec:method}, are trained using the SAR images for each training split, where 1,500 ships and 1,500 vehicles images are subsampled from the visible images, prepared as previously outlined. The network architecture used in this experiment is shown in~Figure~\ref{fig:net_archs}, which follows a standard residual generative network, and the discriminator function uses Spectral Normalization on the convolutional layers. The network training parameters are: \mbox{$\lambda_{\text{s}} = \lambda_{\text{t}} = 10.0$}, \mbox{$\lambda_{\text{cyc}} = 1.0$}, \mbox{$\lambda_{\text{gp}} = 0.01$}, batch \mbox{size $B = 32$}, and number of \mbox{critics $= 2$}, 187,500 training iterations and optimised with Adam~\cite{kingma2015adam} (initial learning \mbox{rate $\eta= 0.0001$}, \mbox{$\beta_1 = 0.5$}, \mbox{$\beta_2 = 0.999$}).

\begin{figure}[tb]
  \centering
    \begin{tabular}{c}
      \begin{minipage}{1.0\hsize}
      \centering
        \includegraphics[clip,width=8.4cm]{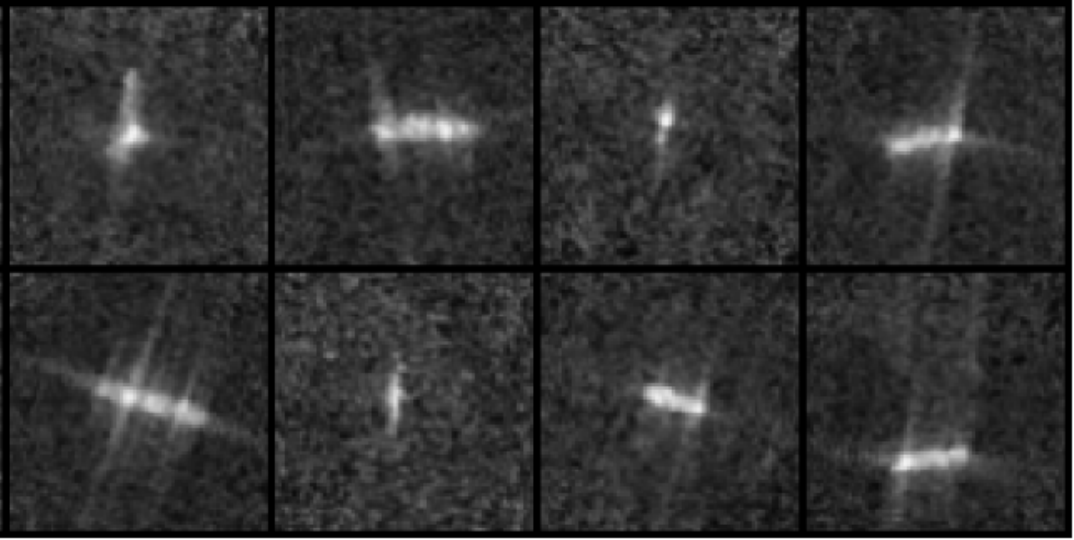}
        \subcaption{Ships}
      \end{minipage} \\
      \begin{minipage}{0.02\hsize}
        \vspace{1mm}
      \end{minipage} \\
      \begin{minipage}{1.0\hsize}
      \centering
        \includegraphics[clip,width=8.4cm]{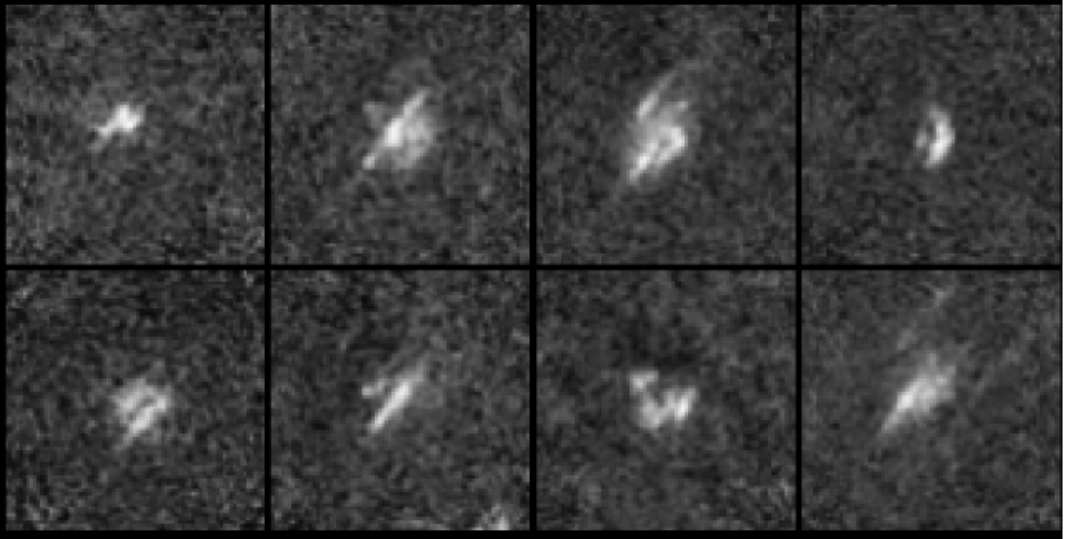}
        \subcaption{Icebergs}
      \end{minipage} \\
      \begin{minipage}{0.02\hsize}
        \vspace{1mm}
      \end{minipage} \\
      \begin{minipage}{1.0\hsize}
      \centering
        \includegraphics[clip,width=8.8cm]{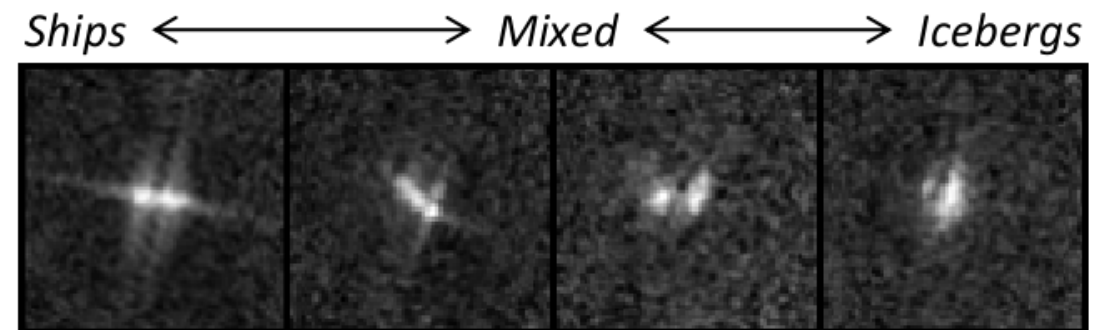}
        \subcaption{Mixed}
      \end{minipage}
    \end{tabular}
    \caption{Examples of the generated SAR images (Train \#1): (a) and (b) are the individual class images. (c) are the inter-class images sorted by the class labels from ship to iceberg.}
    \label{fig:fake_sar}
\end{figure} 

\vspace{3mm}
\subsection{Data Augmentation}
\begin{figure}[tb]
    \centering
        \begin{tabular}{c}
    
          \begin{minipage}{0.9\hsize}
            \centering
              \includegraphics[keepaspectratio, scale=0.6, angle=0]{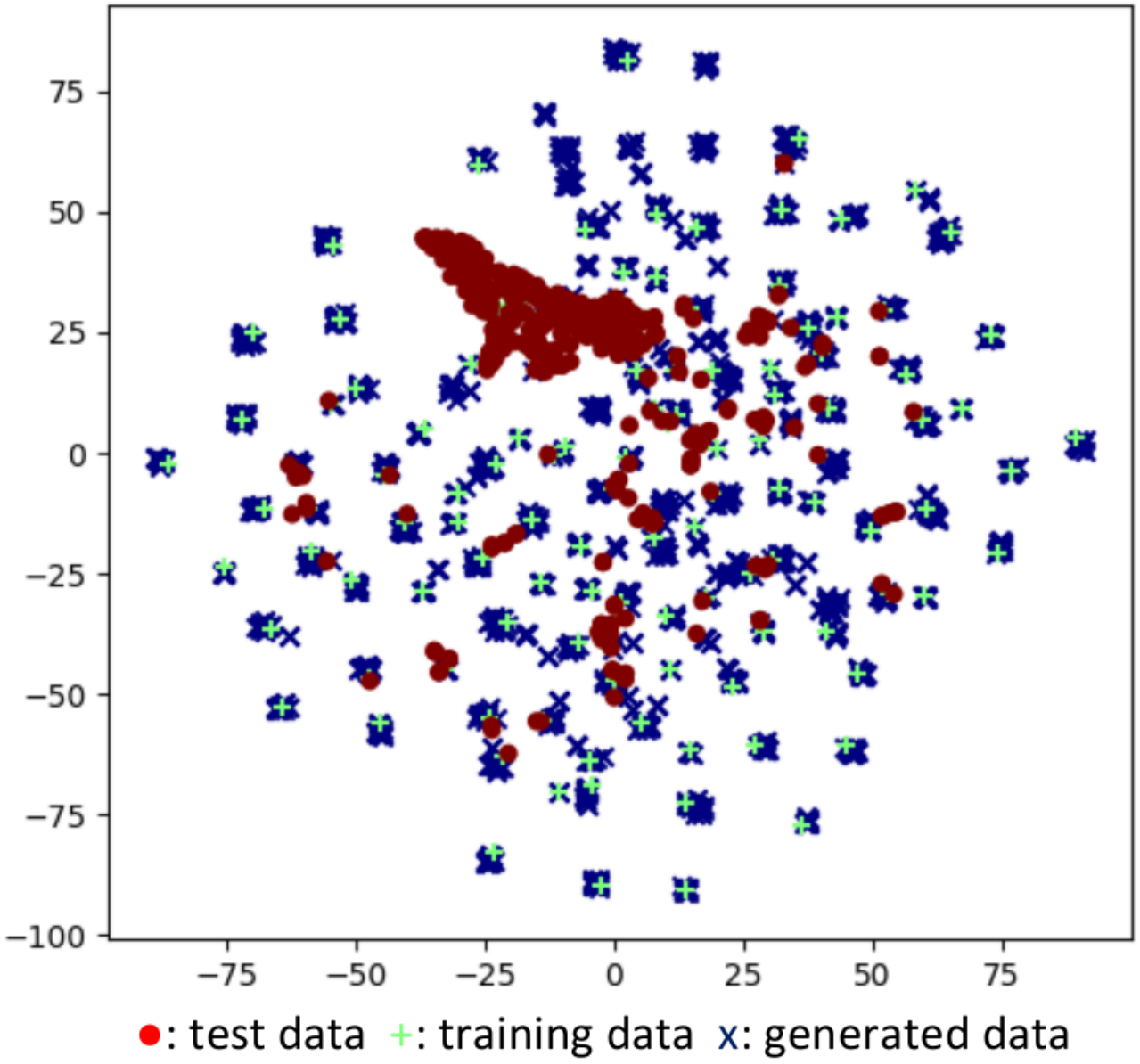}
          \end{minipage} \\
          \begin{minipage}{0.02\hsize}
        \vspace{1mm}
      \end{minipage} \\
          \begin{minipage}{0.9\hsize}
            \centering
              \includegraphics[keepaspectratio, scale=0.6, angle=0]{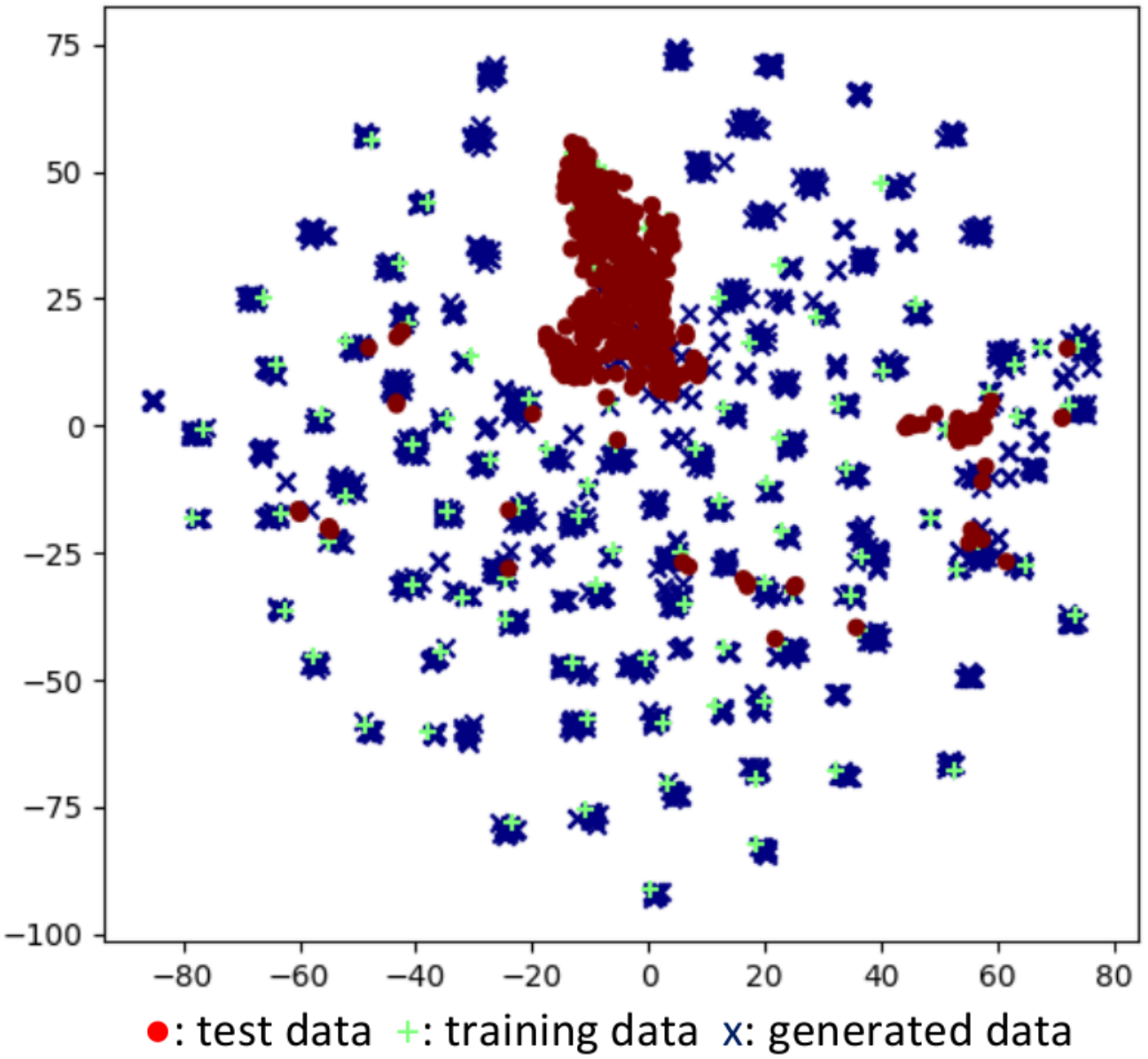}
          \end{minipage}
        \end{tabular}
        \caption{t-SNE plot of ship (top) and iceberg (bottom) images from the test, training and generated datasets (Train \#1).}
        \label{fig:tsne_by_classes}
\end{figure}

Fake SAR images are synthesised using the visible images as the input of our transfer model, as discussed. This results in 3,000 generated SAR images, where examples of these generated images are shown in Figure \ref{fig:fake_sar}. Additionally, we plot the real SAR images and fake SAR images using t-SNE~\cite{maaten2008visualizing} (Figure~\ref{fig:tsne_by_classes}) to show how the different distributions interrelate.
This plot shows that the fake SAR images are well-distributed around the real SAR images.

\vspace{3mm}
\subsection{Evaluation on Object Classification Task}
Evaluation of the classifier performance uses the simple Alexnet architecture~\cite{krizhevsky2012imagenet}, where the classifier performance is compared under the following conditions:
\begin{description}
  \addtolength{\itemindent}{1em}
  \item[$\bullet$]BL: Only using the original training data \cite{statoil-c-core}
  \item[$\bullet$]ROT: BL + rotated 90, 180, and 270 degrees
  \item[$\bullet$]MIXUP: Mixup ($\alpha=0.2$) \cite{zhang2018mixup}
  \item[$\bullet$]MIXCG: BL + MixCycleGAN~\cite{liang2018understanding} \mbox{($\alpha$$=$$0.2$)}
  \item[$\bullet$]C2GMA: (Ours) BL + C2GMA \mbox{($\alpha$$=$$0.2$}, Section \ref{sec:method})
\end{description}
The MixCycleGAN model in this experiment is trained with the same training parameters as our method uses.

The classifiers are trained with the three training datasets, as denoted in Table \ref{tab:dataset}, where the hyperparameters are optimised with the Stochastic Gradient Descent algorithm ($ \eta = 0.02$, number of \mbox{epochs $= 200$}, \mbox{$B = 512$}). Performance is assessed via the testing dataset also outlined in Table ~\ref{tab:dataset}, using statistical accuracy (A), precision (P), recall (R) and F1-score (F1) (Table \ref{tab:accuracy}).
\begin{table*}[tb]
  \begin{center}
    \caption{Overall classification results:  accuracy (A), precision (P), recall (R), and F1-score (F1) on the common test set for each of training sets \#1--3.}
    \begin{minipage}{1.0\hsize}
        \centering
        \begin{tabular}{c  c c c c  c  c c c c  c  c c c c} \hline
          & \multicolumn{4}{c}{Train \#1} && \multicolumn{4}{c}{Train \#2} && \multicolumn{4}{c}{Train \#3} \\ \cline{2-5} \cline{7-10} \cline{12-15}
          & A & P & R & F1 && A & P & R & F1 && A & P & R & F1 \\ \hline \hline
          BL & 0.715 & 0.746 & 0.725 & 0.735 && 0.469 & 0.469 & 0.500 & 0.484 && 0.469 & 0.469 & 0.500 & 0.484 \\
          ROT & 0.707 & 0.723 & 0.714 & 0.719 && 0.469 & 0.469 & 0.500 & 0.484 && 0.469 & 0.469 & 0.500 & 0.484 \\
          MIXUP & 0.766 & 0.794 & 0.775 & 0.784 && 0.690 & 0.728 & 0.701 & 0.714 && 0.690 & 0.694 & 0.681 & 0.688 \\
          MIXCG & 0.760 & 0.765 & 0.764 & 0.765 && 0.757 & 0.783 & 0.766 & 0.776 && 0.676 & 0.708 & 0.687 & 0.697 \\
          C2GMA (Ours) & \bf{0.800} & \bf{0.807} & \bf{0.804} & \bf{0.806} && \bf{0.771} & \bf{0.795} & \bf{0.779} & \bf{0.787} && \bf{0.691} & \bf{0.729} & \bf{0.703} & \bf{0.716} \\ \hline
        \end{tabular}
    \end{minipage}
    \begin{minipage}{0.02\hsize}
        \hspace{2mm}
      \end{minipage}
    \begin{minipage}{1.0\hsize}
        \centering
        \begin{tabular}{c  c c c c} \hline
          & \multicolumn{4}{c}{Average} \\ \cline{2-5}
           & A & P & R & F1 \\ \hline \hline
          BL & 0.551 $\pm$ 0.142 & 0.562 $\pm$ 0.160 & 0.575 $\pm$ 0.130 & 0.568 $\pm$ 0.145 \\
          ROT & 0.549 $\pm$ 0.137 & 0.554 $\pm$ 0.146 & 0.571 $\pm$ 0.124 & 0.562 $\pm$ 0.135 \\
          MIXUP & 0.715 $\pm$ 0.044 & 0.739 $\pm$ 0.051 & 0.719 $\pm$ 0.049 & 0.729 $\pm$ 0.050 \\
          MIXCG & 0.730 $\pm$ 0.048 & 0.752 $\pm$ 0.039 & 0.739 $\pm$ 0.045 & 0.745 $\pm$ 0.042 \\
          C2GMA (Ours) & \bf{0.754} $\pm$ 0.056 & \bf{0.777} $\pm$ 0.042 & \bf{0.762} $\pm$ 0.053 & \bf{0.769} $\pm$ 0.047 \\ \hline
        \end{tabular}
    \end{minipage}
    \label{tab:accuracy}
  \end{center}
\end{table*}

\begin{figure*}[tb]
  \centering
    \begin{tabular}{c}

    \begin{minipage}{0.01\hsize}
        \centering
        \rotatebox[origin=c]{90}{Train \#1}        
      \end{minipage}

      \begin{minipage}{0.15\hsize}
        \centering
          \hspace{6mm} \textnormal{BL}
          \includegraphics[keepaspectratio, scale=0.38, angle=0]{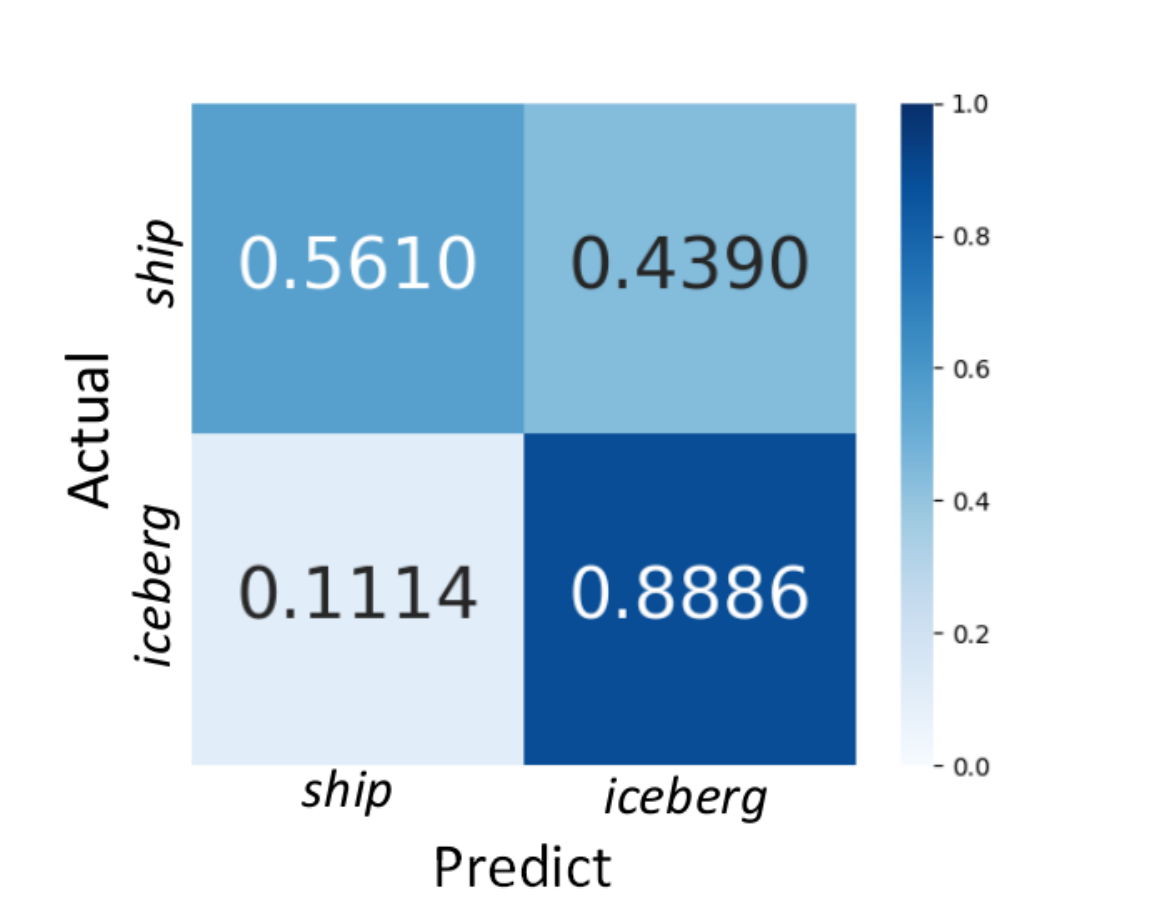}
                          
      \end{minipage}

      \begin{minipage}{0.02\hsize}
        \hspace{2mm}
      \end{minipage}

      \begin{minipage}{0.15\hsize}
        \centering
            \hspace{6mm} \textnormal{ROT}
          \includegraphics[keepaspectratio, scale=0.38, angle=0]{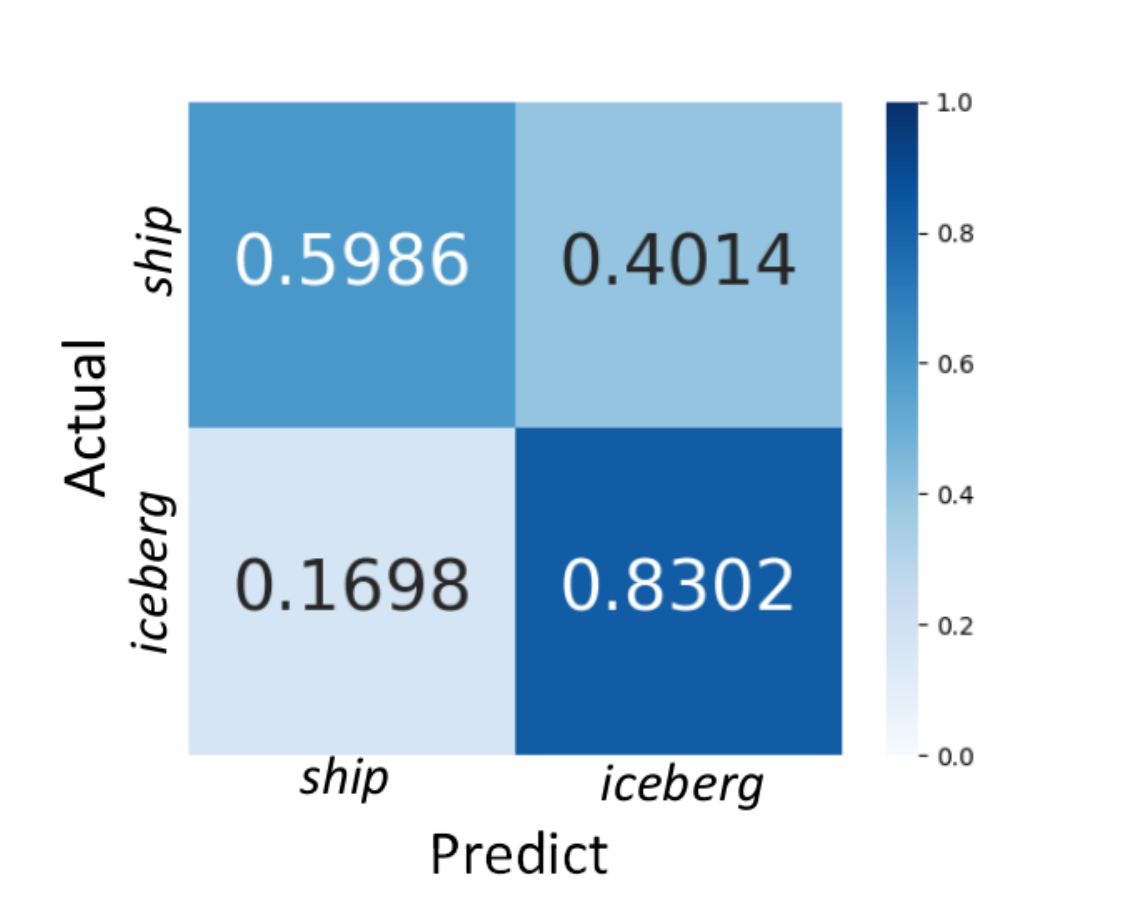}
                          
      \end{minipage}

      \begin{minipage}{0.02\hsize}
        \hspace{2mm}
      \end{minipage}

      \begin{minipage}{0.15\hsize}
        \centering
            \hspace{6mm} \textnormal{MIXUP}
          \includegraphics[keepaspectratio, scale=0.38, angle=0]{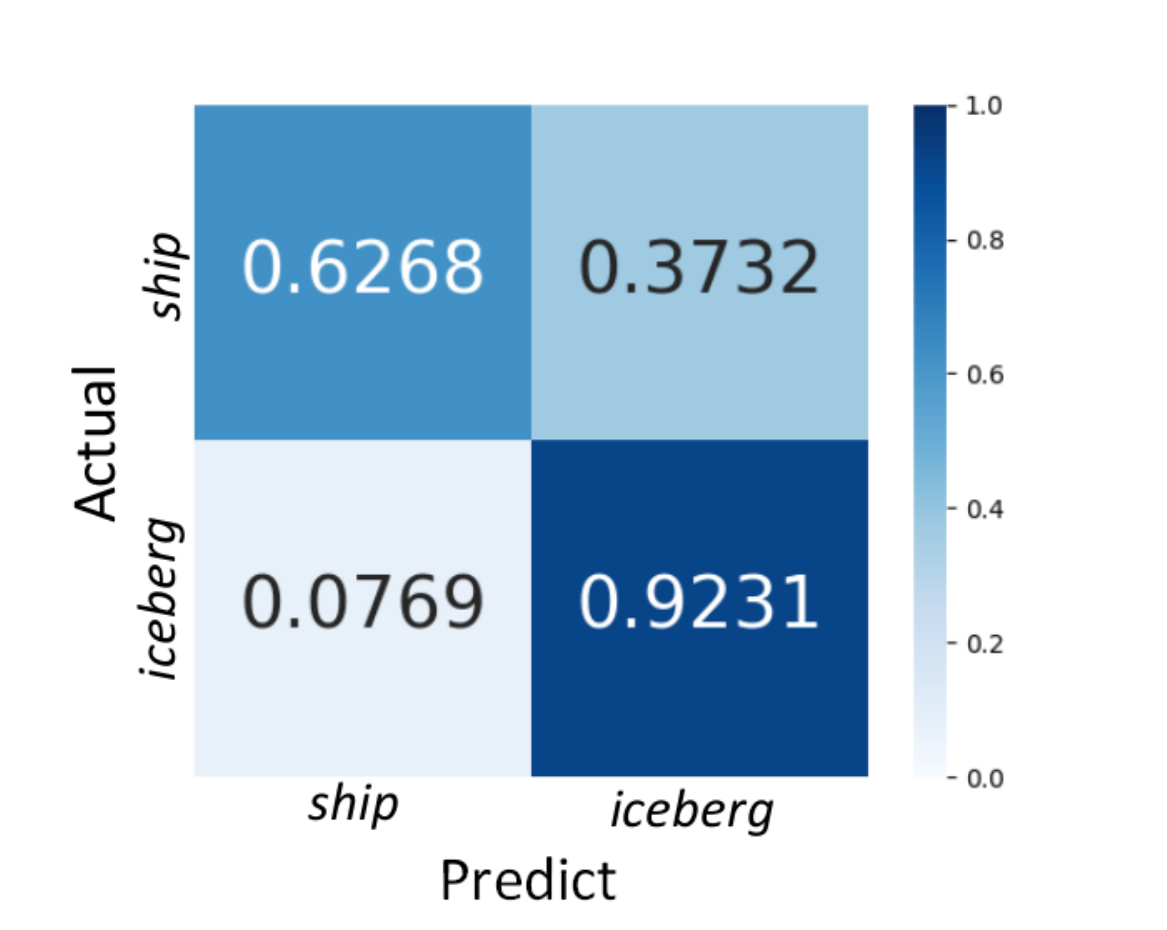}
                          
      \end{minipage}

      \begin{minipage}{0.02\hsize}
        \hspace{2mm}
      \end{minipage}

      \begin{minipage}{0.15\hsize}
        \centering
            \hspace{6mm} \textnormal{MIXCG}
          \includegraphics[keepaspectratio, scale=0.38, angle=0]{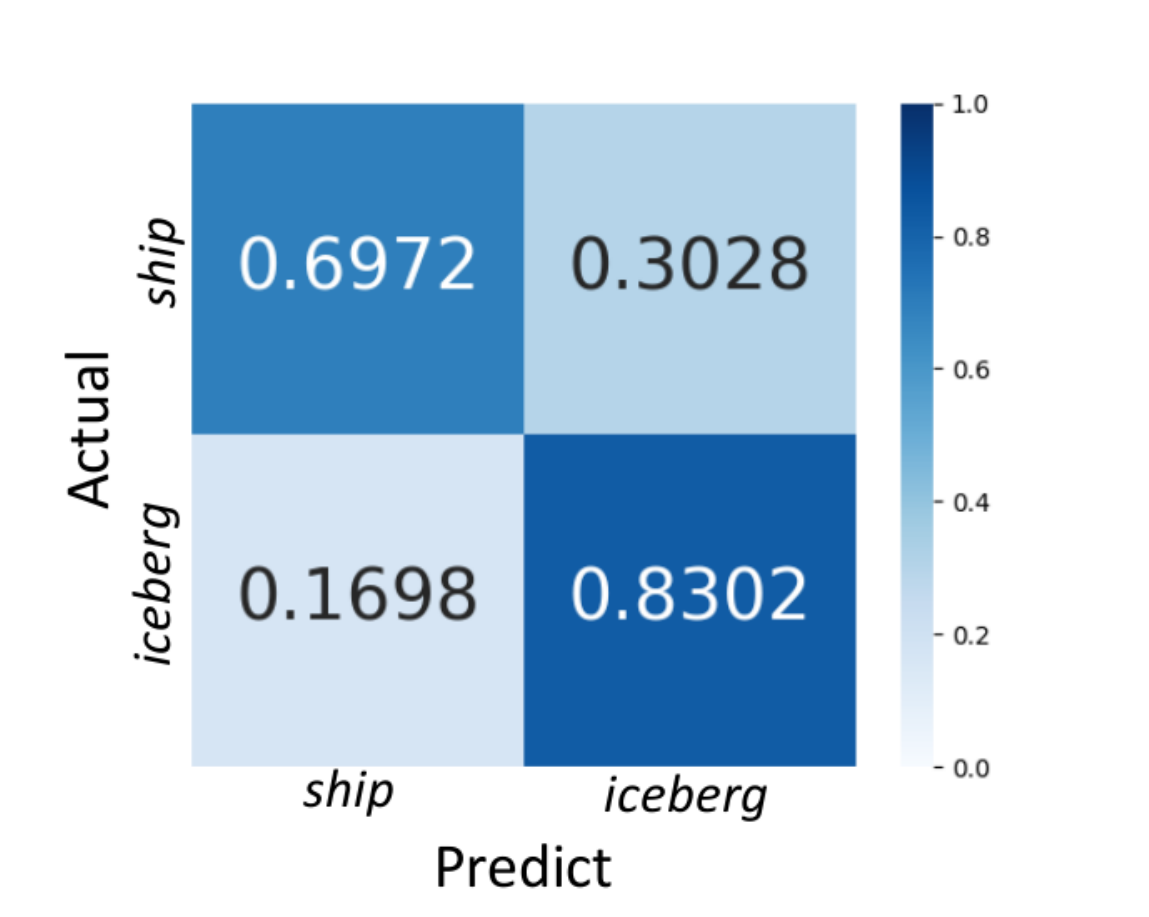}
                                                    
      \end{minipage}

      \begin{minipage}{0.02\hsize}
        \hspace{2mm}
      \end{minipage}

      \begin{minipage}{0.15\hsize}
        \centering
            \hspace{6mm} \textnormal{\small C2GMA (Ours)}
          \includegraphics[keepaspectratio, scale=0.38, angle=0]{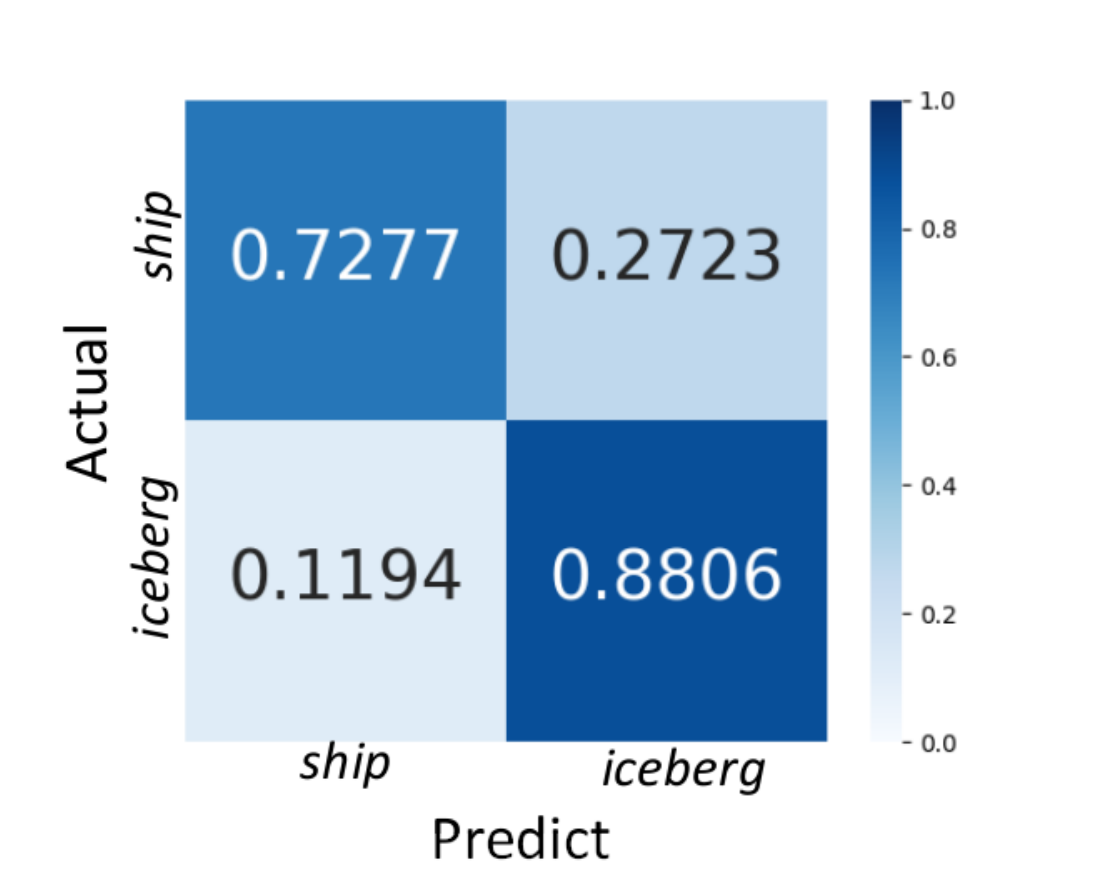}
                          
      \end{minipage} \\

    \begin{minipage}{0.01\hsize}
        \centering
        \rotatebox[origin=c]{90}{Train \#2}        
      \end{minipage}
      
      \begin{minipage}{0.15\hsize}
        \centering
          \includegraphics[keepaspectratio, scale=0.38, angle=0]{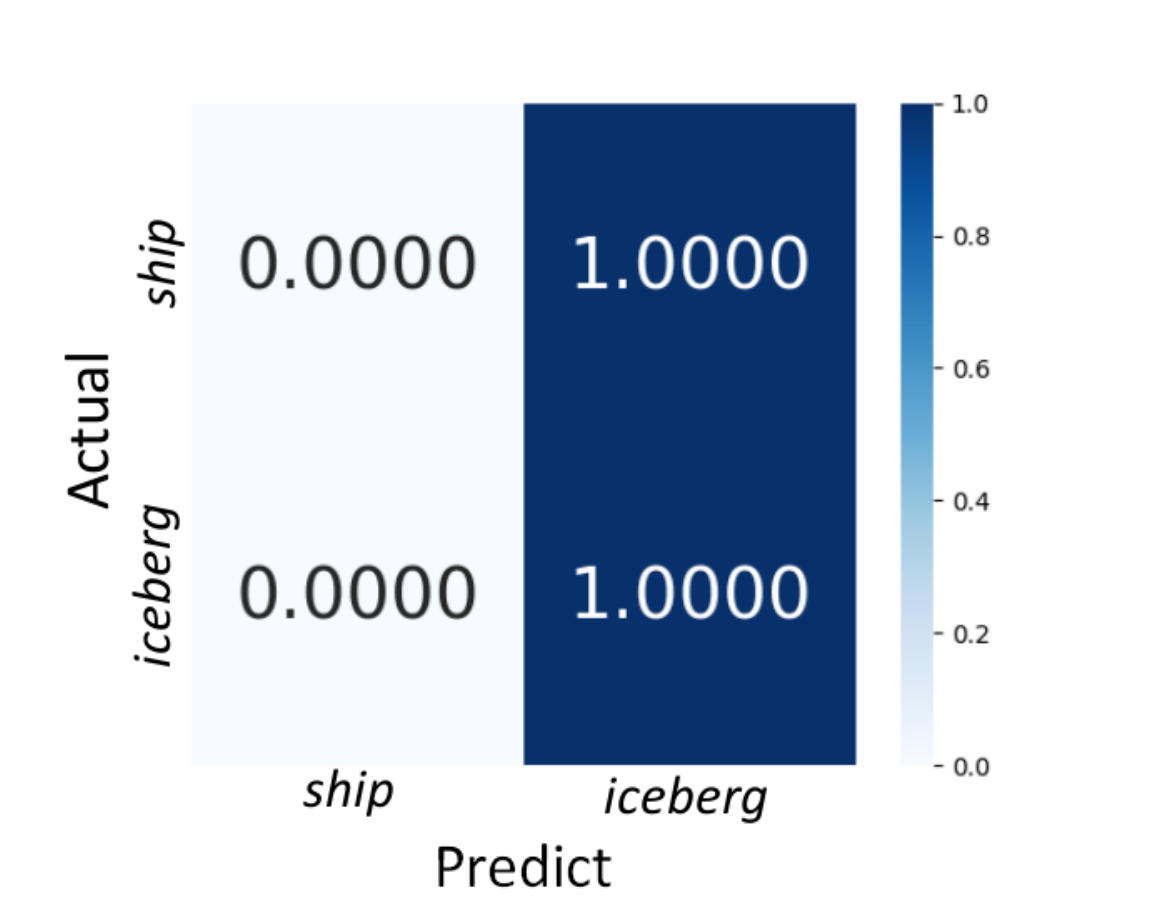}
      \end{minipage}

      \begin{minipage}{0.02\hsize}
        \hspace{2mm}
      \end{minipage}

      \begin{minipage}{0.15\hsize}
        \centering
          \includegraphics[keepaspectratio, scale=0.38, angle=0]{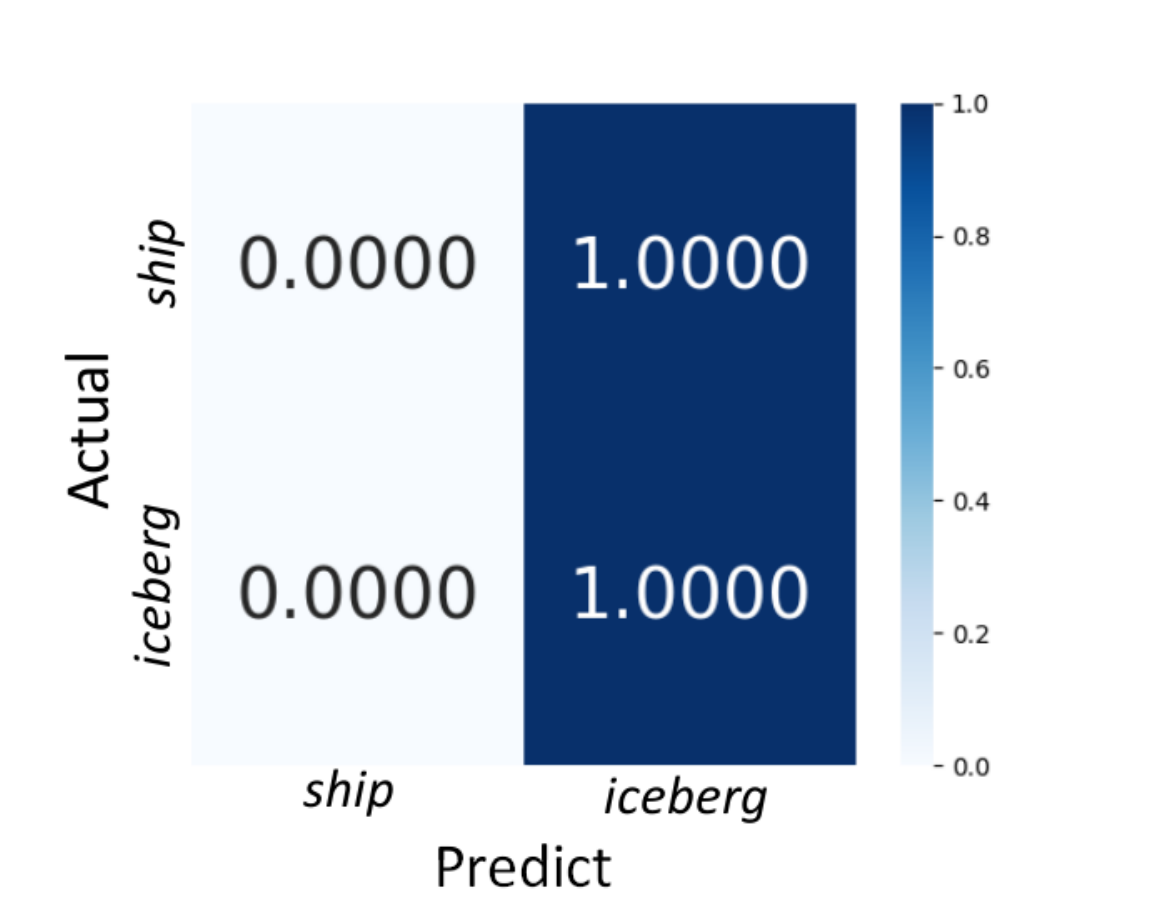}
      \end{minipage}

      \begin{minipage}{0.02\hsize}
        \hspace{2mm}
      \end{minipage}

      \begin{minipage}{0.15\hsize}
        \centering
          \includegraphics[keepaspectratio, scale=0.38, angle=0]{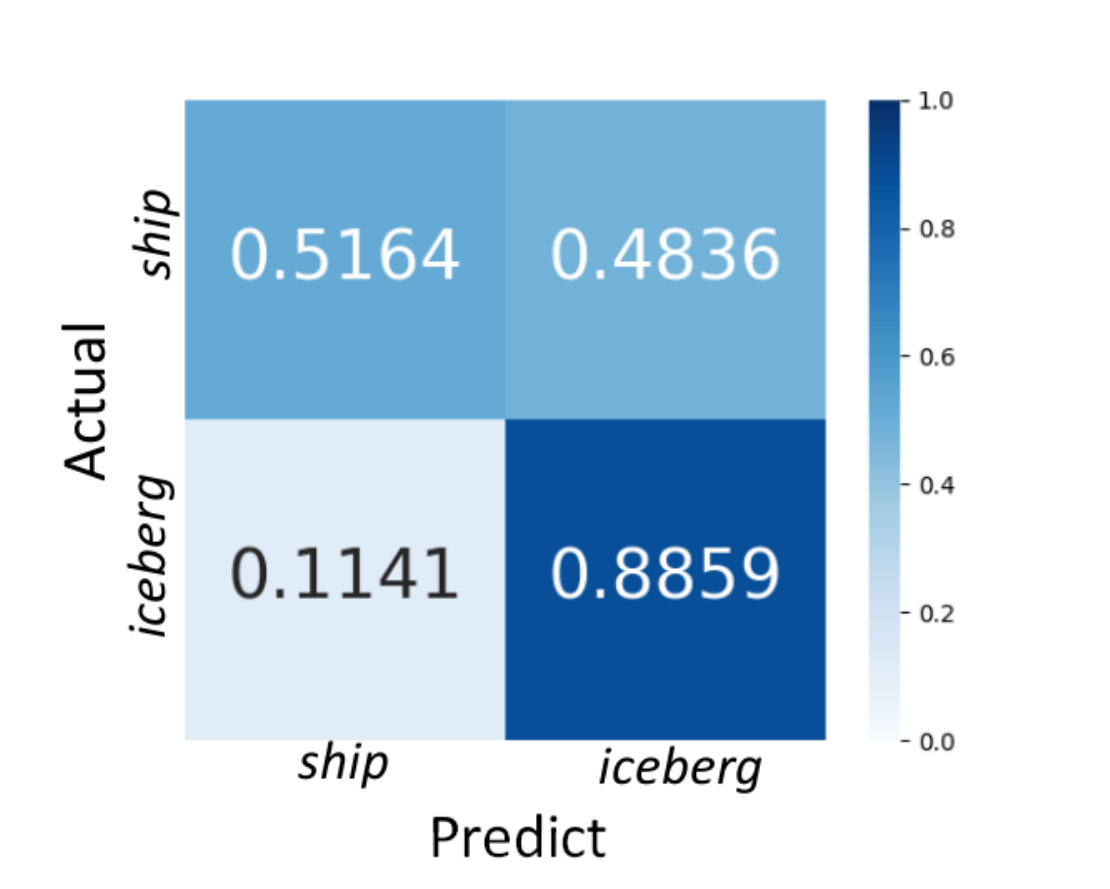}
      \end{minipage}

      \begin{minipage}{0.02\hsize}
        \hspace{2mm}
      \end{minipage}

      \begin{minipage}{0.15\hsize}
        \centering
          \includegraphics[keepaspectratio, scale=0.38, angle=0]{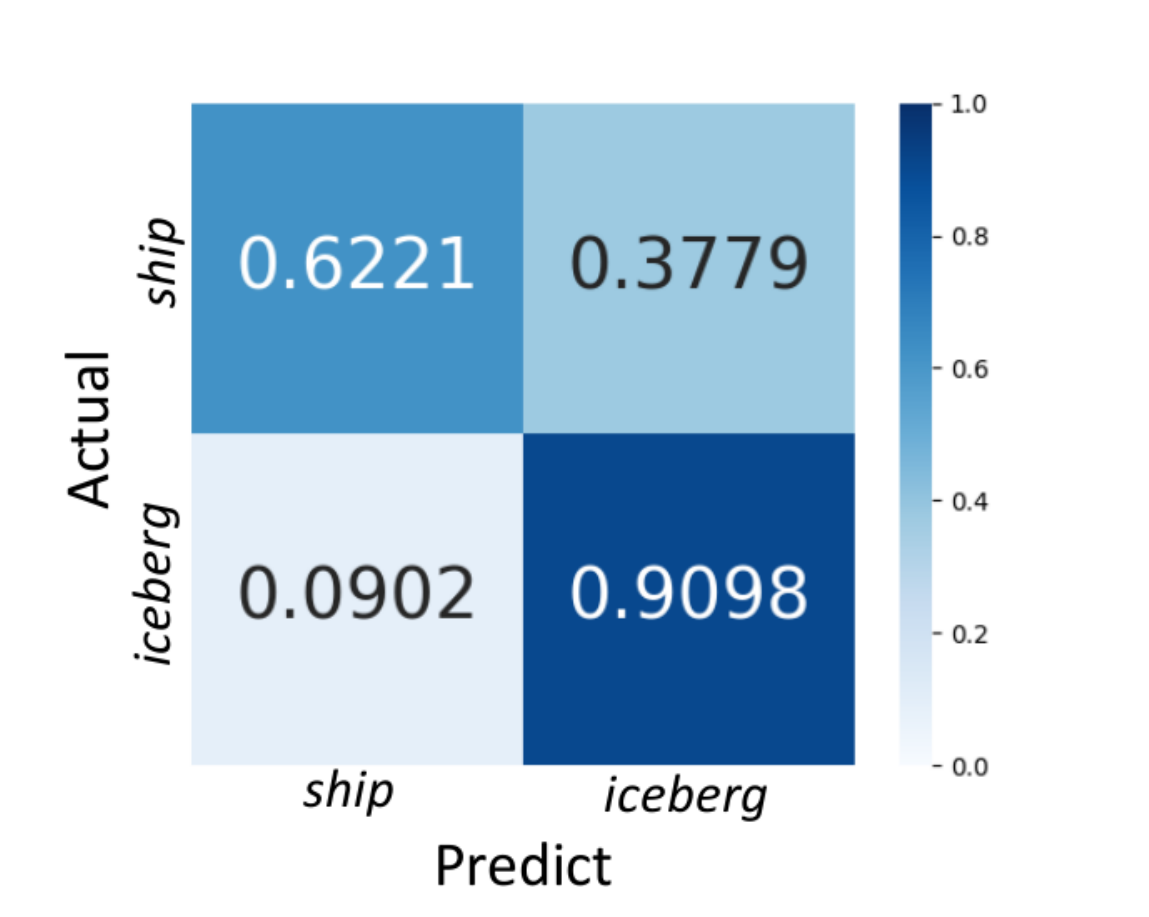}
      \end{minipage}

      \begin{minipage}{0.02\hsize}
        \hspace{2mm}
      \end{minipage}

      \begin{minipage}{0.15\hsize}
        \centering
          \includegraphics[keepaspectratio, scale=0.38, angle=0]{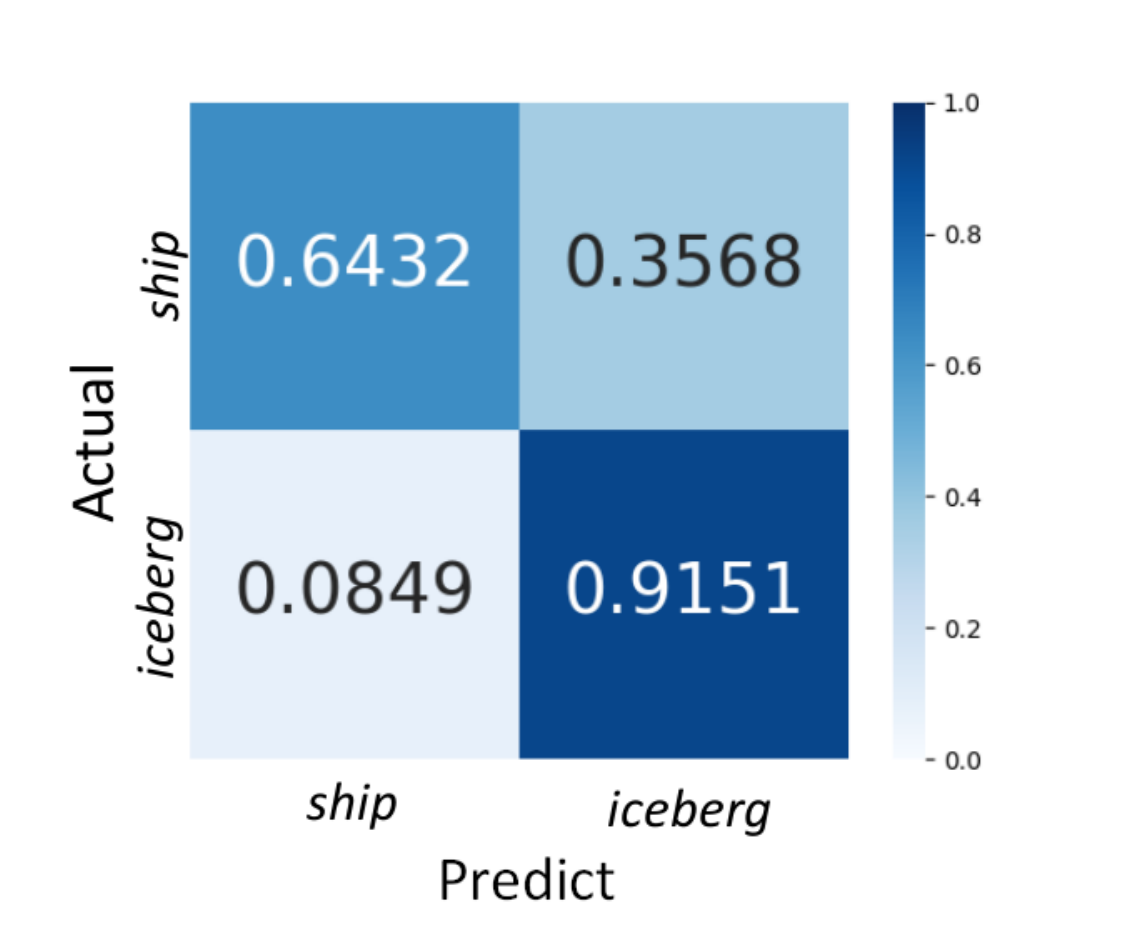}
      \end{minipage} \\

    \begin{minipage}{0.01\hsize}
        \centering
        \rotatebox[origin=c]{90}{Train \#3}        
      \end{minipage}

      \begin{minipage}{0.15\hsize}
        \centering
          \includegraphics[keepaspectratio, scale=0.38, angle=0]{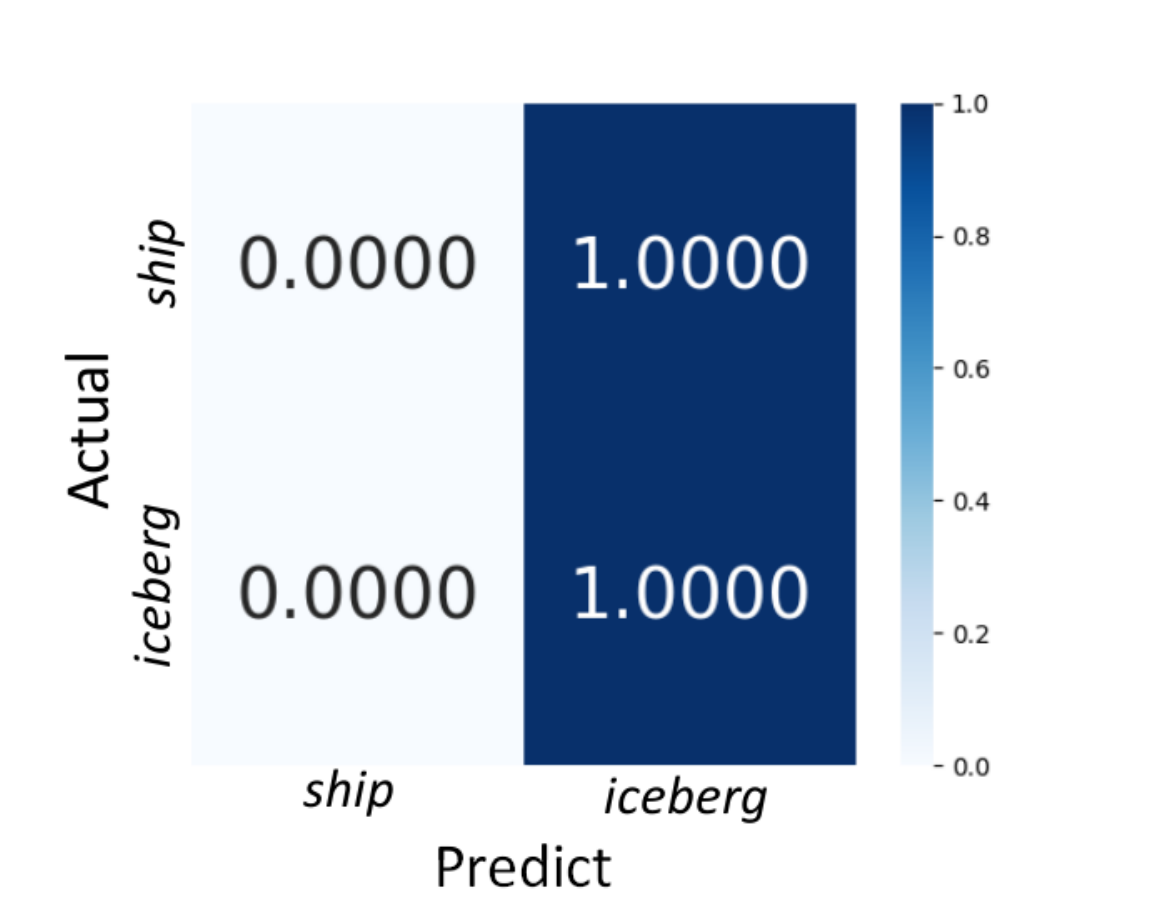}
      \end{minipage}

      \begin{minipage}{0.02\hsize}
        \hspace{2mm}
      \end{minipage}

      \begin{minipage}{0.15\hsize}
        \centering
          \includegraphics[keepaspectratio, scale=0.38, angle=0]{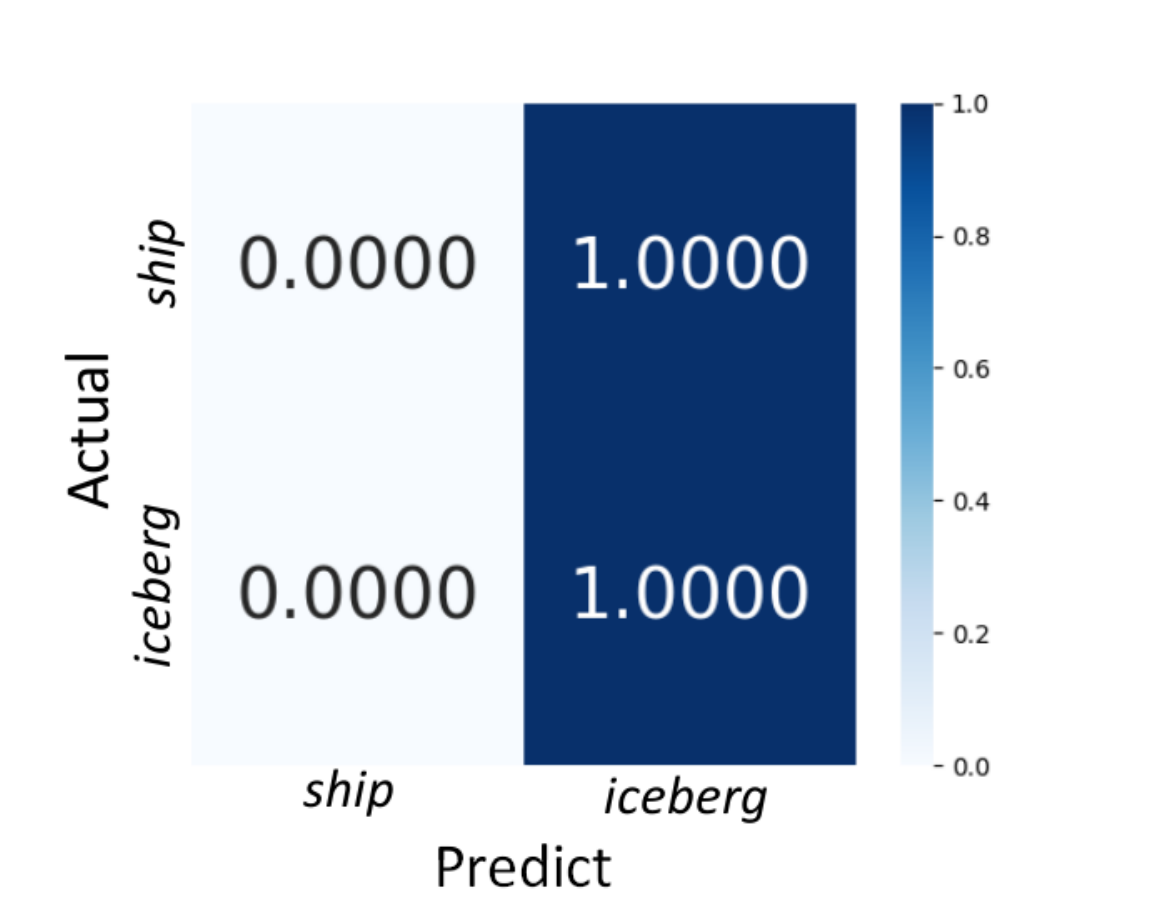}
      \end{minipage}

      \begin{minipage}{0.02\hsize}
        \hspace{2mm}
      \end{minipage}

      \begin{minipage}{0.15\hsize}
        \centering
          \includegraphics[keepaspectratio, scale=0.38, angle=0]{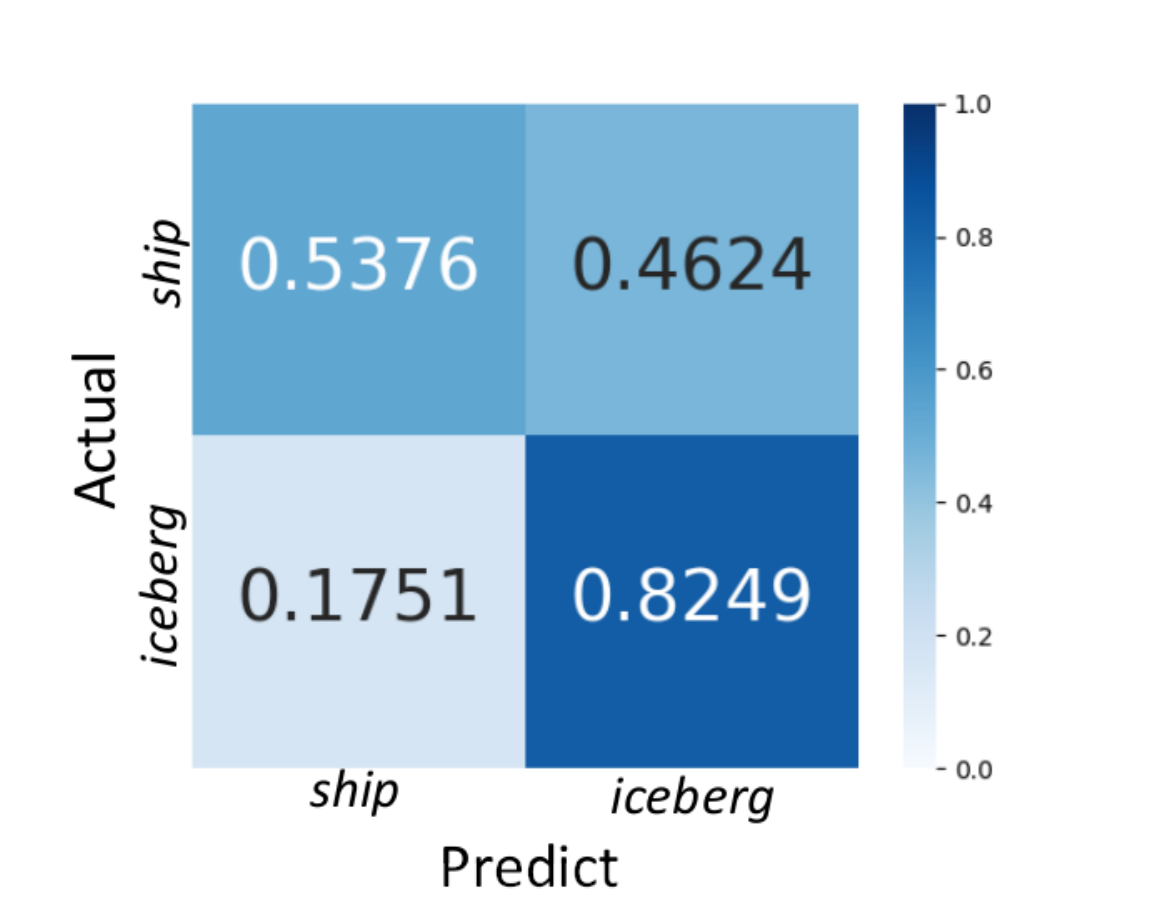}
      \end{minipage}

      \begin{minipage}{0.02\hsize}
        \hspace{2mm}
      \end{minipage}

      \begin{minipage}{0.15\hsize}
        \centering
          \includegraphics[keepaspectratio, scale=0.38, angle=0]{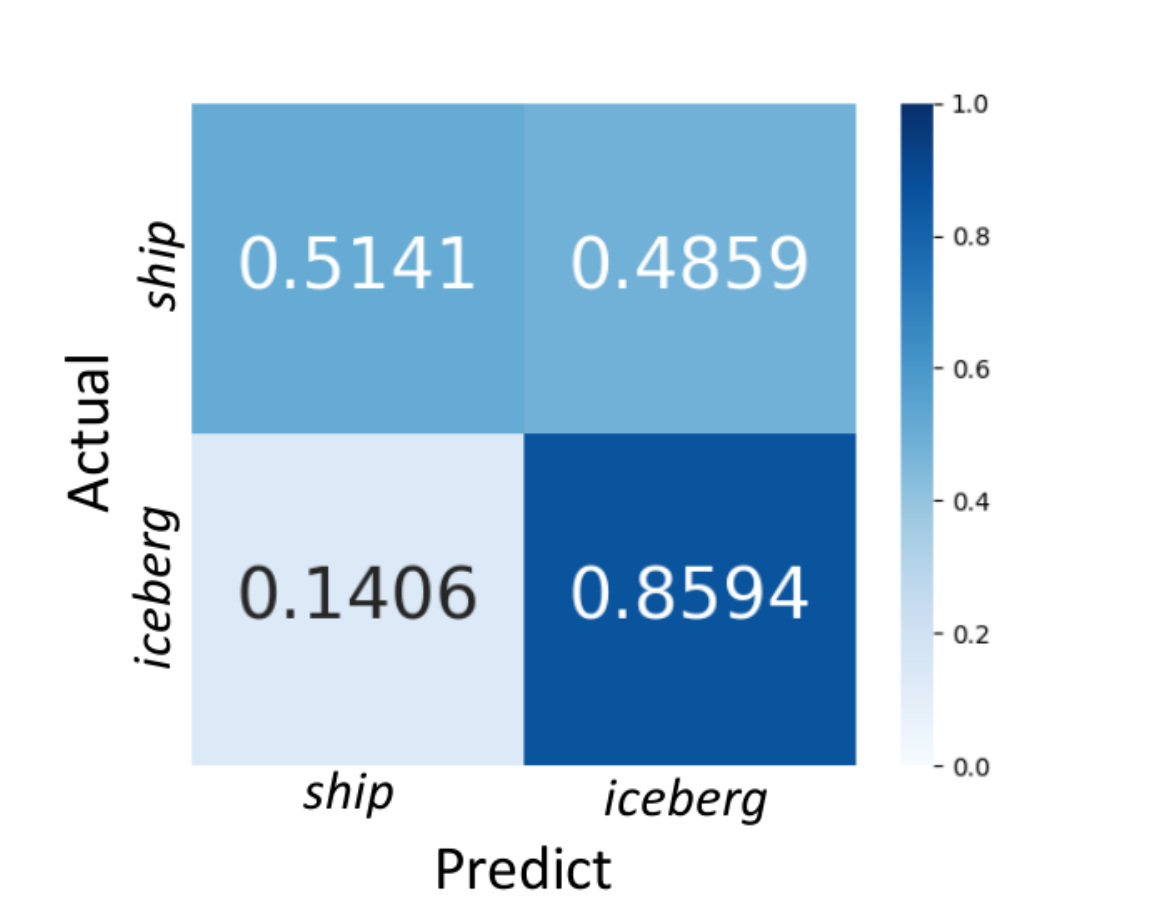}
      \end{minipage}

      \begin{minipage}{0.02\hsize}
        \hspace{2mm}
      \end{minipage}

      \begin{minipage}{0.15\hsize}
        \centering
          \includegraphics[keepaspectratio, scale=0.38, angle=0]{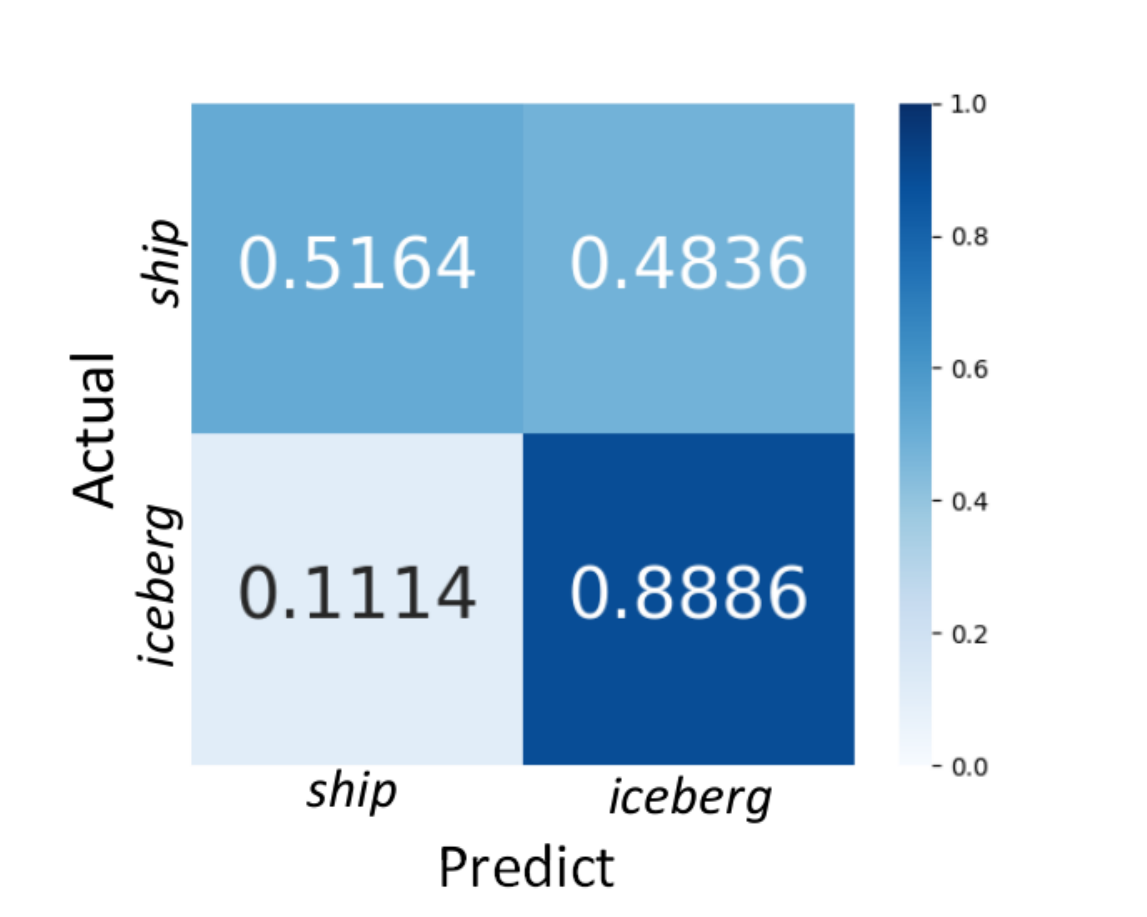}
      \end{minipage} \\

    \end{tabular}
    \caption{Per-class performance (confusion matrices) of our approach (C2GMA) against prior work in the field.}
    \label{fig:cm}
\end{figure*}

Quantitative results are shown in Table \ref{tab:accuracy}, alongside the additional individual per-class classification performances for ships and icebergs, shown in the confusion matrices in Figure~\ref{fig:cm}.
The overall results show that our proposed C2GMA data augmentation approach significantly outperforms the other approaches (BL, ROT, MIXUP~\cite{zhang2018mixup}, and MIXCG~\cite{liang2018understanding}). We find that generating new images using our approach increases training data appropriately, where the process of synthesising inter-class images is shown to provide significant improvements for the overall classification performance (C2GMA, Table~\ref{tab:accuracy}).


\section{Conclusion}
\label{sec:conclusion}

This paper proposes and evaluates a CycleGAN enabled data augmentation approach, Conditional CycleGAN Mixup Augmentation (C2GMA), to address the challenge of effective data augmentation within cross-domain imagery where the availability of one of the domains is limited.
In particular, we show that the generation of interpolated mixed class (non-visible domain) image examples via our novel C2GMA methodology leads to a significant improvement in the quality of non-visible domain classification tasks that suffer due to limited data availability and variety. 
Focusing on classification within the synthetic aperture radar domain, our approach is evaluated on a variation of the Statoil/C-CORE Iceberg Classifier Challenge dataset and achieves 75.4\% accuracy, demonstrating a significant improvement when compared against traditional augmentation strategies. 
Future work will consider DNN architecture modifications to enable generation of higher quality images for improved classification results and applications to other non-visible band imaging domains.


\bibliographystyle{IEEEtran}
\bibliography{references}


\end{document}